\documentclass[twoside,11pt]{article}

\usepackage{jmlr2e}

\usepackage{amsmath}
\usepackage{amsfonts}
\usepackage{nicefrac}
\usepackage{microtype}
\usepackage{multirow}
\usepackage{array}
\usepackage{booktabs}
\usepackage{tabularx}
\usepackage{xcolor}
\usepackage{colortbl}

\colorlet{cTraj}{orange}
\colorlet{cRew}{green!55!black}
\colorlet{cBoth}{red!75!black}
\colorlet{cAgent}{purple}
\colorlet{cOPD}{teal}

\colorlet{tabheadTraj}{cTraj!22}
\colorlet{tabbandTraj}{cTraj!10}
\colorlet{tabheadRew}{cRew!15}
\colorlet{tabbandRew}{cRew!7}
\colorlet{tabheadBoth}{cBoth!13}
\colorlet{tabbandBoth}{cBoth!6}
\colorlet{tabheadAG}{cAgent!16}    
\colorlet{tabbandAG}{cAgent!7}     
\colorlet{tabheadOPD}{cOPD!18}     
\colorlet{tabbandOPD}{cOPD!8}      
\colorlet{tabheadI}{blue!12}       

\newcommand{\cmark}{\textcolor{green!55!black}{\checkmark}}
\newcommand{\pmark}{\textcolor{orange!85!black}{$\circ$}}
\newcommand{\xmark}{\textcolor{black!35}{\texttimes}}
\usepackage{tikz}
\usetikzlibrary{arrows.meta, positioning, shapes.geometric, fit, backgrounds, calc}
\usepackage[normalem]{ulem}

\hypersetup{hidelinks,pdfborder={0 0 0}}

\newcommand{\E}{\mathbb{E}}

\newcommand{\pitheta}{\pi_{\theta}}
\newcommand{\piold}{\pi_{\theta_{\mathrm{old}}}}
\newcommand{\piref}{\pi_{\mathrm{ref}}}
\newcommand{\KL}{\mathrm{KL}}
\newcommand{\clip}{\mathrm{clip}}

\usepackage{lastpage}

\makeatletter
\def\ps@jmlrtps{\let\@mkboth\@gobbletwo
  \def\@oddhead{\scriptsize Preprint \hfill}%
  \def\@oddfoot{\hfil\small\rm\thepage\hfil}%
  \def\@evenhead{\scriptsize Preprint \hfill}%
  \def\@evenfoot{\hfil\small\rm\thepage\hfil}}
\long\def\blfootnote#1{%
  \begingroup
    \long\def\@makefntext##1{\parindent 1em\noindent ##1}%
    \renewcommand\@makefnmark{}%
    \footnotetext{#1}%
  \endgroup
}
\def\@starteditor{}\def\@endeditor{}
\makeatother

\ShortHeadings{A First-Principles Derivation of LLM Policy Optimization}{A First-Principles Derivation of LLM Policy Optimization}
\firstpageno{1}

\begin{document}

\title{A First-Principles Derivation of LLM Policy Optimization:
       From Expected Reward to GRPO
       and Its Structural Extensions}

\author{%
  \normalsize
  \begin{minipage}{\textwidth}
  \centering
  \def\authorsep{\hskip 1.0em plus 0.4em minus 0.2em}
  \mbox{Jianghan Shen\textsuperscript{1,2,$*$}}\authorsep
  \mbox{Siqi Luo\textsuperscript{2,3,$*$}}\authorsep
  \mbox{Yue Li\textsuperscript{4}}\authorsep
  \mbox{Jiyao Liu\textsuperscript{2,5}}\authorsep
  \mbox{Wanying Qu\textsuperscript{2,5}}\\[3pt]
  \mbox{Yi Zhang\textsuperscript{6}}\authorsep
  \mbox{Ziyan Huang\textsuperscript{2}}\authorsep
  \mbox{Tianbin Li\textsuperscript{2}}\authorsep
  \mbox{Ming Hu\textsuperscript{2}}\\[3pt]
  \mbox{Xiaohong Liu\textsuperscript{3,7,$\dagger$}}\authorsep
  \mbox{Yirong Chen\textsuperscript{2,$\dagger$}}\authorsep
  \mbox{Junjun He\textsuperscript{7,$\dagger$}}
  \\[8pt]
  \small\rm
  \textsuperscript{1}Nanjing University \quad
  \textsuperscript{2}Shanghai Artificial Intelligence Laboratory \quad
  \textsuperscript{3}Shanghai Jiao Tong University \\[2pt]
  \textsuperscript{4}Peking University \quad
  \textsuperscript{5}Fudan University \quad
  \textsuperscript{6}Nanjing University of Aeronautics and Astronautics \\[2pt]
  \textsuperscript{7}Shanghai Innovation Institute
  \end{minipage}
}

\editor{}

\maketitle
\thispagestyle{jmlrtps}   

\blfootnote{\textsuperscript{$*$}These authors contributed equally.\quad
  \textsuperscript{$\dagger$}Corresponding authors.\\
  Correspondence: \texttt{jianghanshencs@smail.nju.edu.cn}}

\begin{abstract}
%
Policy gradient algorithms for language models optimize the same objective $J(\theta) = \E_{\tau \sim p_\theta(\tau)}[R(\tau)] $,
which has exactly two factors: the trajectory probability $p_\theta(\tau)$ and the reward $R(\tau)$.
Every method from REINFORCE to PPO to GRPO and their descendants modifies one or both factors to address a specific failure in the preceding formulation.
Existing surveys organize these methods by domain or chronology, which obscures the rationale behind each design choice and the precise location of its intervention within the gradient estimator.
%
%
This survey revisits the landscape of LLM policy optimization from $J(\theta)$ on \textbf{first principles} and uses the trajectory side, induced by $p_\theta(\tau)$, and the reward side, induced by $R(\tau)$, as the two axes along which methods are located.
It covers the path from REINFORCE and PPO to GRPO, as well as post-GRPO variants, Agentic RL, and GRPO-OPD.
The resulting framework is \textbf{unified}, \textbf{diagnostic}, and \textbf{extensible}: it analyzes methods from a shared objective, identifies which side each method modifies and why, and applies the same trajectory and reward axes across these settings.
Across these settings, the framework also exposes compound failures that no single-side fix resolves and that therefore require joint design of the trajectory side and the reward side.
The boundary cases and coupled failures identified by this map mark where existing solutions run out and provide a principled starting point for designing the next generation of LLM policy optimization algorithms.

\end{abstract}


\section{Introduction}
\label{sec:intro}

Reinforcement learning has become a central mechanism for post-training large language models toward human preferences, task intent, and reasoning behavior~\citep{christiano2017deep}.
Early language-model \textbf{RLHF} (\textbf{R}einforcement \textbf{L}earning from \textbf{H}uman \textbf{F}eedback) methods applied preference-trained reward models to continuation and summarization tasks~\citep{ziegler2019fine,stiennon2020learning}.
\textbf{PPO}-based (\textbf{P}roximal \textbf{P}olicy \textbf{O}ptimization) RLHF then established the standard optimization template for instruction following and assistant alignment~\citep{schulman2017proximal,ouyang2022training,bai2022training}.
\textbf{GRPO} (\textbf{G}roup \textbf{R}elative \textbf{P}olicy \textbf{O}ptimization) marked a turning point for reasoning-oriented RL by replacing the learned critic with group-relative reward normalization, thereby reducing the memory burden of PPO-style training while enabling strong mathematical and reasoning performance~\citep{shao2024deepseekmath,guo2025deepseek}.
After GRPO, the field expanded rapidly in several directions.
Post-GRPO variants modify how rollouts are collected and how training signals are constructed~\citep{yu2026dapo,lightman2024let}.
Agentic RL extends policy optimization to multi-turn interaction with tools and external feedback~\citep{jin2025search,wei2026swe}.
\textbf{OPD} (\textbf{O}n-\textbf{P}olicy \textbf{D}istillation) trains a student against a teacher signal on its own rollouts. In its original form it replaces $J(\theta)$ with a divergence-based objective~\citep{gu2024minillm}, which exits the policy-gradient frame and is treated as a boundary case in Section~\ref{sec:alternatives}. This survey instead studies the \textbf{GRPO-OPD hybrid}, which incorporates the teacher signal into the GRPO update while retaining $J(\theta)$ as the optimization objective~\citep{ramos2026combining}.
\textbf{DPO} (\textbf{D}irect \textbf{P}reference \textbf{O}ptimization) and related methods constitute another line of work.
This line replaces the policy-gradient objective with a preference-based objective derived directly from preference data~\citep{rafailov2023direct}.
Together, these methods define a research landscape that is broad, rapidly evolving, and difficult to navigate.

Existing surveys~\citep{kaufmann2023survey,wang2023aligning,gao2024towards,wang2024reinforcement,wang2024comprehensive,liu2025reinforcement,srivastava2025technical,zhang2025landscape,song2026survey} provide broad coverage of this landscape, but they typically organize methods by pipeline stage, method family, or application domain.
Such organizations are useful for coverage, but they do not answer the diagnostic question practitioners face: \emph{given a failure, what mechanism produced it, and what minimal intervention can resolve it without introducing a worse failure elsewhere?}
Answering this question requires tracing each method back to the optimization objective it modifies.
For policy-gradient methods, this objective is $J(\theta)=\E_{\tau\sim p_\theta(\tau)}[R(\tau)]$, whose update is governed by two factors: the trajectory probability $p_\theta(\tau)$ and the reward $R(\tau)$.
These two factors naturally define the trajectory side and the reward side of policy optimization.
Failures can therefore be diagnosed by asking whether they arise from how trajectories are sampled, reused, or constrained, from how rewards are assigned and normalized, or from the interaction between the two.
A few recent surveys propose unified formulations~\citep{wang2024comprehensive,song2026survey}, but they mainly unify the form of existing methods rather than use these two factors to explain failure origins and intervention choices.
This survey addresses this gap by organizing every reviewed method as a transition from a concrete failure mode to a corresponding intervention on the trajectory side, the reward side, or both.
Table~\ref{tab:survey_comparison} summarizes how this differs from existing work.

\begin{table}[h]
\centering
\caption{Comparison with related surveys on LLM policy optimization.
\emph{From $J(\theta)$}: starts from $J(\theta)=\E[R(\tau)]$ and its two factors;
\emph{Diagnostic}: locates each method by the factor it modifies and the failure it answers;
\emph{Agentic}/\emph{OPD}: whether the two axes carry to agentic RL and on-policy distillation.
\cmark~full, \pmark~partial, \xmark~none.
}

\label{tab:survey_comparison}
\small
\setlength{\tabcolsep}{5pt}
\renewcommand{\arraystretch}{1.1}
\begin{tabularx}{\textwidth}{>{\raggedright\arraybackslash}p{4.1cm}>{\raggedright\arraybackslash}Xcccc}
\toprule
\rowcolor{tabheadI}
Survey & Organizing principle & From $J(\theta)$ & Diagnostic & Agentic & OPD \\
\midrule
\citet{wang2023aligning}          & Alignment: data, training, evaluation & \xmark & \xmark & \xmark & \xmark \\
\citet{gao2024towards}            & Preference learning, four components  & \pmark    & \xmark & \xmark & \xmark \\
\citet{wang2024reinforcement}     & Feedback type: RLHF, RLAIF, DPO       & \xmark & \xmark & \xmark & \xmark \\
\citet{kaufmann2023survey}        & RLHF pipeline stages                  & \xmark & \xmark & \xmark & \xmark \\
\citet{srivastava2025technical}   & Algorithm families and applications   & \xmark & \xmark & \xmark & \xmark \\
\citet{liu2025reinforcement}      & LLM lifecycle stages                  & \xmark & \xmark & \xmark & \xmark \\
\citet{song2026survey}            & OPD: $f$-divergence, design axes      & \xmark & \xmark & \pmark    & \cmark \\
\citet{zhang2025landscape}        & Agentic capabilities and domains      & \xmark & \xmark & \cmark & \xmark \\
\citet{wang2024comprehensive}     & Unified policy-gradient estimator     & \pmark    & \xmark & \pmark    & \xmark \\
\midrule
\rowcolor{blue!6}
\textbf{This survey} & First-principles two-axis diagnostic map & \cmark & \cmark & \cmark & \cmark \\
\bottomrule
\end{tabularx}
\end{table}

The organizing principle behind these distinctions is the core objective itself.
All policy-gradient algorithms for language models optimize a trajectory-level objective,
\[
  J(\theta) \;=\; \E_{\tau \sim p_\theta(\tau)}\bigl[R(\tau)\bigr]
\]
where $\tau$ denotes a sampled trajectory, $p_\theta(\tau)$ denotes the probability of that trajectory under the current policy, and $R(\tau)$ denotes the reward assigned to it.
This objective exposes two natural axes of intervention: the trajectory side, which controls how samples enter the update, and the reward side, which controls what signal weights that update.
From REINFORCE~\citep{williams1992simple} to PPO~\citep{schulman2017proximal} and GRPO~\citep{shao2024deepseekmath}, methods can be understood as modifying one or both axes to address failures left unresolved by the preceding formulation.
This decomposition turns $J(\theta)$ into a \textbf{diagnostic map}: \emph{given any failure, the question becomes which factor it originates from and what targeted modification resolves it.}

Applying this map to the post-GRPO landscape reveals three classes of methods.
\uline{Trajectory-side methods} ask which trajectories should be used and how much each should be optimized, spanning rollout collection, compression, diversity, repair, and the clipping and ratio rules that govern the update~\citep{zheng2025gspo,zheng2026greso}.
\uline{Reward-side methods} ask how to accurately evaluate the quality of each policy decision, and act at two levels. At the reward signal $R_i$, they cover density, source, shaping, and multi-objective aggregation. At the advantage $\hat{A}_i$ that converts $R_i$ into a per-token weight, they cover statistical bias correction and fine-grained credit assignment~\citep{lightman2024let,cui2025process,liu2025understanding}.
\uline{Both-side methods} address failure modes that couple the two components, where modifying one side alone leaves residual instability on the other~\citep{yu2026dapo,yang2025dcpo}.

The same two-axis map also extends beyond the post-GRPO single-turn setting. %
\textbf{Agentic RL} expands the trajectory side to multi-turn environment interaction, but this expansion also introduces reward-side challenges such as delayed feedback and temporal credit assignment~\citep{jin2025search,wei2026swe,feng2025retool}.
\textbf{The GRPO-OPD hybrid} expands the reward side with dense teacher signals, but these signals depend on trajectory-side properties such as rollout quality and update stability~\citep{ramos2026combining,xu2025kdrl}.
Thus, the map identifies open problems not only within each side, but also where trajectory construction and reward construction must be optimized jointly.

This survey analyzes existing methods under a first-principles two-axis framework.
The framework has three properties.
\textbf{(i) unified and first-principles:} all methods are revisited from the shared objective $J(\theta)=\E[R(\tau)]$ rather than treated as isolated algorithm families.
\textbf{(ii) diagnostic:} each method is located by which side it modifies, what failure motivates the modification, and what residual instability remains.
\textbf{(iii) extensible: }the same trajectory and reward axes apply beyond single-turn GRPO to Agentic RL and the \textbf{GRPO-OPD} (\textbf{G}roup \textbf{R}elative \textbf{P}olicy \textbf{O}ptimization with \textbf{O}n-\textbf{P}olicy \textbf{D}istillation) hybrid, revealing why these settings require joint design of the trajectory side and the reward side.
By contrast, methods that replace $J(\theta)$ itself, such as DPO~\citep{rafailov2023direct} and divergence-minimization distillation~\citep{gu2024minillm,agarwal2023gkd}, fall outside this policy-gradient frame and are discussed separately as boundary cases.

\begin{figure}[t]
\centering
\includegraphics[width=\textwidth]{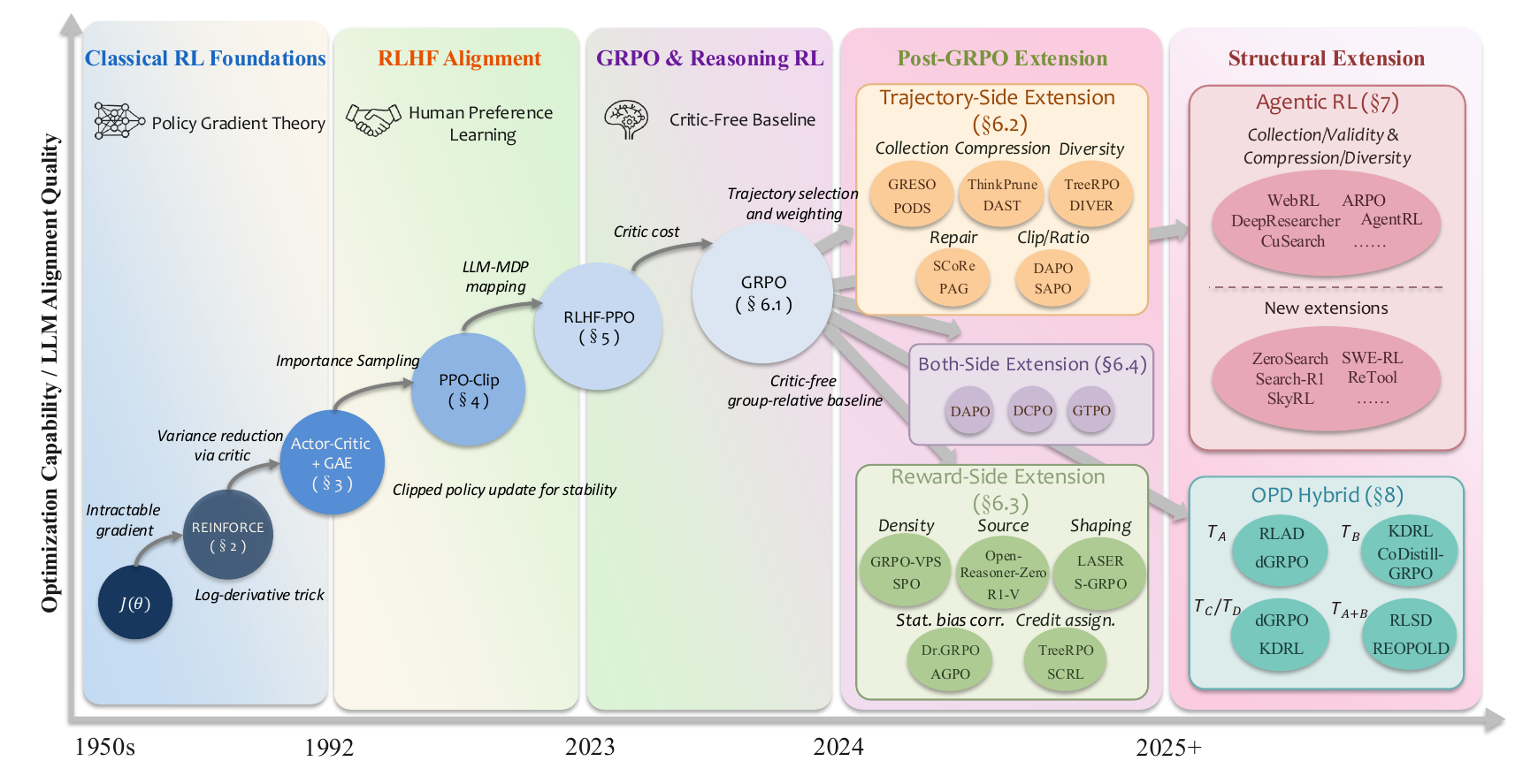}
\caption{%
A first-principles landscape of LLM policy optimization. The blue spine runs from $J(\theta)=\mathbb{E}[R(\tau)]$ to GRPO, each arrow labeling the failure that motivates the next step. Post-GRPO variants split into the trajectory side (\S\ref{sec:traj_variants}), the reward side (\S\ref{sec:reward_variants}), and both-side instabilities (\S\ref{sec:both_variants}). Agentic RL (\S\ref{sec:agentic}) extends the trajectory side to multi-turn interaction, and the GRPO-OPD hybrid (\S\ref{sec:opd_line}) extends the reward side through four operators: $\mathcal{T}_A$ importance ratio, $\mathcal{T}_B$ advantage, $\mathcal{T}_C$ in-expectation distillation, $\mathcal{T}_D$ external regularizer, with $\mathcal{T}_{A+B}$ coupling the first two. Each group shows representative methods only.
}
\label{fig:chain}
\end{figure}

The survey covers online policy optimization for autoregressive LLMs.
\textbf{Part I} starts from $J(\theta)=\E[R(\tau)]$ and develops the trajectory and reward axes through REINFORCE, Actor-Critic, GAE, PPO, and the LLM--MDP mapping.
\textbf{Part II} uses these axes to organize post-GRPO variants into trajectory-side, reward-side, and coupled interventions.
\textbf{Part III} extends the same diagnostic framework to Agentic RL and the GRPO-OPD hybrid.
Section~\ref{sec:discussion} marks the boundary of the policy-gradient frame and identifies open problems.
Figure~\ref{fig:chain} provides an overview of the first-principles diagnostic map.
Together, the three parts show that the rapid post-GRPO expansion is not a collection of independent proposals but a structured response to failures that arise, in a predictable order, along the trajectory and reward axes of $J(\theta)$.

\part*{Part~I\quad From the RL Objective to the Per-Step LLM Form}
\addcontentsline{toc}{part}{Part I: From the RL Objective to the Per-Step LLM Form}

\section{From Objective to Policy Gradient}
\label{sec:pg_section}

\subsection{The Optimization Objective and Its Two Sides}
\label{sec:mdp}

A Markov Decision Process is a tuple $(\mathcal{S},\mathcal{A},P,R,\rho_0,\gamma)$, where $\mathcal{S}$ is the state space, $\mathcal{A}$ the action space, $P\colon\mathcal{S}\times\mathcal{A}\to\Delta(\mathcal{S})$ the transition kernel, $R\colon\mathcal{S}\times\mathcal{A}\to\mathbb{R}$ the reward function, $\rho_0$ the initial state distribution, and $\gamma\in[0,1]$ the discount factor~\citep{sutton2018reinforcement}.
The defining property is the Markov condition~\citep{sutton2018reinforcement}: transitions depend only on the current state and action,
\begin{equation}
  P(s_{t+1}\mid s_0,a_0,\ldots,s_t,a_t) = P(s_{t+1}\mid s_t,a_t).
\end{equation}
A trajectory $\tau=(s_0,a_0,r_0,s_1,a_1,r_1,\ldots,s_T)$ is generated by sampling $s_0\sim\rho_0$, selecting $a_t\sim\pitheta(\cdot\mid s_t)$ at each step, receiving $r_t=R(s_t,a_t)$, and transitioning to $s_{t+1}\sim P(\cdot\mid s_t,a_t)$.
By the Markov property the trajectory probability factorizes as
\begin{equation}
  \label{eq:traj_prob}
  p_\theta(\tau)
  = \rho_0(s_0)
    \prod_{t=0}^{T-1}\pitheta(a_t\mid s_t)\,P(s_{t+1}\mid s_t,a_t),
\end{equation}
and the discounted return is $R(\tau)=\sum_{t=0}^{T-1}\gamma^t r_t$~\citep{sutton2018reinforcement}.

\paragraph{The core objective.}
All methods in this survey optimize
\begin{equation}
  \label{eq:objective}
  J(\theta)
  = \E_{\tau\sim p_\theta(\tau)}[R(\tau)]
  = \sum_\tau p_\theta(\tau)\,R(\tau).
\end{equation}
This objective weights each trajectory by two factors: how likely the current policy makes it, and how much reward it yields.
Both factors must be addressed for gradient optimization to be effective, and each introduces a distinct challenge~\citep{williams1992simple,sutton1999policy}.

\textbf{Trajectory side.}
Computing $\nabla_\theta J(\theta)$ requires differentiating through $p_\theta(\tau)$.
Summing over all trajectories is intractable, and the naive gradient
\begin{equation}
\label{eq:naive_grad}
\nabla_\theta J(\theta) = \sum_\tau R(\tau)\nabla_\theta p_\theta(\tau)
\end{equation}
is not expressed in the form $\E_{p_\theta}[\cdot]$.
It therefore cannot be estimated directly by sampling from the current policy.
Section~\ref{sec:pg} addresses this by applying the log-derivative trick, which converts the gradient into a sampling-friendly expectation.

\textbf{Reward side.}
Even after the gradient becomes tractable, the trajectory return $R(\tau)=\sum_t \gamma^t r_t$ assigns a single scalar to the entire trajectory and uses it to credit all actions uniformly.
This assignment is causally coarse because action $a_t$ cannot influence rewards obtained before time $t$.
The result is high variance and uniform credit across all steps regardless of each action's actual contribution.
Section~\ref{sec:actor_critic_section} addresses this by introducing the advantage function and per-step estimation.

\subsection{The Policy Gradient Theorem}
\label{sec:pg}

\paragraph{Log-derivative trick.}
The identity $\nabla_\theta p_\theta(\tau) = p_\theta(\tau)\nabla_\theta\log p_\theta(\tau)$~\citep{williams1992simple} converts the intractable gradient~\eqref{eq:naive_grad} into a sampling-friendly expectation:
\begin{equation}
  \label{eq:pg_expect}
  \nabla_\theta J(\theta)
  = \E_{\tau\sim p_\theta}\bigl[
      R(\tau)\,\nabla_\theta\log p_\theta(\tau)
    \bigr].
\end{equation}

\paragraph{Eliminating environment terms.}
Differentiating $\log p_\theta(\tau)$ from Equation~\eqref{eq:traj_prob} gives
\begin{equation}
  \nabla_\theta\log p_\theta(\tau)
  = \underbrace{\nabla_\theta\log\rho_0(s_0)}_{=0}
  + \sum_{t=0}^{T-1}\nabla_\theta\log\pitheta(a_t|s_t)
  + \underbrace{\sum_{t=0}^{T-1}\nabla_\theta\log P(s_{t+1}|s_t,a_t)}_{=0}.
\end{equation}
Both the initial-state term and the transition term vanish because they do not depend on $\theta$.
The policy-gradient identity therefore reduces to
\begin{equation}
  \label{eq:reinforce}
  \nabla_\theta J(\theta)
  = \E_{\tau\sim p_\theta}\!\left[
      R(\tau)\sum_{t=0}^{T-1}
      \nabla_\theta\log\pitheta(a_t|s_t)
    \right],
\end{equation}
which gives the REINFORCE estimator when the expectation is approximated by sampled trajectories~\citep{williams1992simple}.
This trajectory-level form is consistent with the policy gradient theorem of \citet{sutton1999policy}.

The log-derivative trick resolves the trajectory-side problem: the gradient is now a sampling-friendly expectation, and transition-dynamics derivatives have been eliminated.
The reward-side problem remains open.

\section{From REINFORCE to Actor-Critic}
\label{sec:actor_critic_section}

The reward-side problem is the coarseness of the credit signal.
REINFORCE weights every action log-probability term by the full trajectory return $R(\tau)$, even though action $a_t$ cannot influence rewards before time $t$, and therefore provides a causally coarse and high-variance credit signal.
This section addresses it in two stages.
First, reward-to-go removes causally irrelevant pre-$t$ rewards from the weight on each action, and the advantage function~\citep{sutton2018reinforcement} replaces absolute return with a relative quality signal.
Second, TD bootstrapping and GAE~\citep{schulman2015gae} make per-step advantage estimation practical without requiring full Monte Carlo rollouts.
The Actor-Critic architecture~\citep{konda1999actor,mnih2016a3c} combines both stages into the first complete per-timestep optimization form.

\subsection{Advantage Estimation}
\label{sec:advantage}

\paragraph{Reward-to-go.}

Replacing the full trajectory return with the causal future return
\begin{equation}
  G_t = \sum_{k=t}^{T-1}\gamma^{k-t}r_k
\end{equation}
removes rewards obtained before time $t$ from the weight applied to $\nabla_\theta\log\pitheta(a_t|s_t)$.

This replacement does not alter the expected gradient.
Let $C_t=\sum_{k=0}^{t-1}\gamma^k r_k$ denote the part of the trajectory return that precedes time $t$.
Conditional on the pre-action history $h_t=(s_0,a_0,r_0,\ldots,s_t)$, $C_t$ is fixed, while $a_t$ is sampled from $\pitheta(\cdot\mid s_t)$.
Therefore, by iterated expectation,
\[
\E\!\left[
  C_t\nabla_\theta\log\pitheta(a_t\mid s_t)
\right]
=
\E\!\left[
  C_t\,
  \E_{a_t\sim\pitheta(\cdot\mid s_t)}
  [\nabla_\theta\log\pitheta(a_t\mid s_t)]
\right].
\]
The inner expectation is zero because
\[
\E_{a_t\sim\pitheta}[\nabla_\theta\log\pitheta(a_t\mid s_t)]
=
\sum_a \pitheta(a\mid s_t)\nabla_\theta\log\pitheta(a\mid s_t)
=
\sum_a \nabla_\theta \pitheta(a\mid s_t)
=
\nabla_\theta 1
=
0.
\]
Thus, rewards before time $t$ contribute zero in expectation.
Removing them changes the variance but not the expected gradient~\citep{sutton2018reinforcement,greensmith2004variance}.

The resulting reward-to-go policy gradient is
\begin{equation}
  \label{eq:reward_to_go}
  \nabla_\theta J(\theta)
  = \E\!\left[
      \sum_{t=0}^{T-1}
      G_t\,\nabla_\theta\log\pitheta(a_t\mid s_t)
    \right].
\end{equation}

\paragraph{Advantage function.}
Although $G_t$ removes rewards obtained before time $t$, it still measures the absolute future return from the trajectory.
For policy optimization, however, the relevant question is not only whether the return after $s_t$ was large, but whether the sampled action $a_t$ produced a return larger than what is typically achievable from the same state.
This motivates comparing two conditional expectations of the return.
The first is the average return obtained from state $s_t$ under the policy,
\begin{equation}
  V^\pi(s_t)
  =
  \E_\pi[G_t \mid s_t],
\end{equation}
which serves as a state-dependent baseline.
The second is the average return obtained when the first action is fixed to $a_t$ and the policy is followed thereafter,
\begin{equation}
  Q^\pi(s_t,a_t)
  =
  \E_\pi[G_t \mid s_t,a_t].
\end{equation}
Their difference defines the advantage function~\citep{sutton2018reinforcement}:
\begin{equation}
  A^\pi(s_t,a_t)
  =
  Q^\pi(s_t,a_t) - V^\pi(s_t).
\end{equation}
Thus, $A^\pi(s_t,a_t)$ measures the relative quality of $a_t$ at $s_t$: it is positive when $a_t$ yields higher expected return than the policy's average return from $s_t$, and negative when it yields lower expected return.

\paragraph{TD error.}
The advantage definition above is expressed in terms of expected returns, but computing $G_t$ exactly requires observing a complete episode~\citep{sutton2018reinforcement}.
Temporal-difference learning avoids this requirement by bootstrapping from a learned critic $V_\phi$.
It forms the one-step temporal-difference error
\begin{equation}
  \label{eq:td}
  \delta_t
  =
  r_t + \gamma V_\phi(s_{t+1}) - V_\phi(s_t),
\end{equation}
where $r_t$ and $s_{t+1}$ are drawn from a single sampled transition.
When the critic is exact, $V_\phi = V^\pi$, the Bellman expectation equation gives
\begin{equation}
  Q^\pi(s_t,a_t)
  =
  \E_\pi\!\left[
    r_t + \gamma V^\pi(s_{t+1})
    \mid s_t,a_t
  \right].
\end{equation}
Taking the conditional expectation of $\delta_t$ over the sampled transition,
\begin{align}
  \E_\pi[\delta_t \mid s_t,a_t]
  &=
  \E_\pi\!\left[
    r_t + \gamma V^\pi(s_{t+1}) - V^\pi(s_t)
    \mid s_t,a_t
  \right] \\
  &=
  Q^\pi(s_t,a_t) - V^\pi(s_t) \\
  &=
  A^\pi(s_t,a_t).
\end{align}
Thus, under an exact value function, the expected TD error is an unbiased one-step estimate of the advantage~\citep{konda1999actor}.
A single-sample $\delta_t$ is therefore an unbiased but noisy estimator in this ideal case.
In practice, $V_\phi$ is learned and approximate, so $\delta_t$ inherits the critic's approximation error.
It can still have much lower variance than Monte Carlo estimates based directly on $G_t$, but it may now be biased~\citep{sutton2018reinforcement}.
Monte Carlo returns are unbiased but high-variance, while one-step TD bootstrapping reduces variance at the cost of critic-induced bias when the value function is approximate.
This tension between the two extremes motivates a practical interpolation, which the Actor-Critic architecture provides through GAE.

\subsection{Actor-Critic: First Complete Per-Timestep Form}
\label{sec:actor_critic}

The Actor-Critic architecture~\citep{konda1999actor,mnih2016a3c} maintains two networks: an actor $\pitheta$ that selects actions, and a critic $V_\phi$ that estimates state values for advantage computation.

\paragraph{Generalized Advantage Estimation.}
GAE~\citep{schulman2015gae} interpolates between Monte Carlo and one-step TD estimation by accumulating discounted TD errors,
\begin{equation}
  \label{eq:gae}
  A_t^{\mathrm{GAE}}
  =
  \sum_{l=0}^{T-t-1}(\gamma\lambda)^l\,\delta_{t+l},
\end{equation}
where $\delta_{t+l} = r_{t+l} + \gamma V_\phi(s_{t+l+1}) - V_\phi(s_{t+l})$ denotes the TD error at step $t+l$.
At a true terminal state, the terminal bootstrap value is set to $V_\phi(s_T)=0$.
For truncated rollouts, the final value is instead bootstrapped from the critic.
This sum admits an efficient backward recurrence,
\[
  A_t^{\mathrm{GAE}}
  =
  \delta_t + \gamma\lambda A_{t+1}^{\mathrm{GAE}},
\]
seeded at the horizon with $A_T^{\mathrm{GAE}}=0$, which avoids explicit summation over future steps.
The parameter \(\lambda\in[0,1]\) controls the bias--variance trade-off.
When \(\lambda=0\), GAE reduces to the single-step TD error, producing a low-variance but potentially biased estimate.
When \(\lambda=1\), it accumulates all future TD errors and recovers the Monte Carlo advantage estimator $G_t - V_\phi(s_t)$, which is unbiased but high-variance.
Intermediate values yield a geometrically weighted mixture, with smaller \(\lambda\) prioritizing variance reduction through bootstrapping and larger \(\lambda\) reducing reliance on the critic~\citep{schulman2015gae,sutton2018reinforcement}.

\paragraph{Actor and critic losses.}
The actor is updated using GAE advantages:
\begin{equation}
  L_{\mathrm{actor}}(\theta)
  = -\E_t\!\left[A_t^{\mathrm{GAE}}\log\pitheta(a_t|s_t)\right].
\end{equation}
The critic minimizes the squared error against a bootstrapped return target.
In the single-step case the target is $r_t + \gamma V_\phi(s_{t+1})$, evaluated at the data-collecting critic and treated as a constant with respect to $\phi$.
With GAE the target generalizes to the $\lambda$-return,
\begin{equation}
  \label{eq:lambda_return}
  \hat{R}_t = A_t^{\mathrm{GAE}} + V_t^{\mathrm{old}},
\end{equation}
where $V_t^{\mathrm{old}} = V_{\phi_{\mathrm{old}}}(s_t)$ is the value estimate from the data-collecting critic.
Both $A_t^{\mathrm{GAE}}$ and $V_t^{\mathrm{old}}$ are computed from the data-collecting critic and held fixed during the update of $\phi$.
The critic loss is then
\begin{equation}
  \label{eq:critic_loss}
  L_{\mathrm{critic}}(\phi)
  = \E_t\!\left[\bigl(V_\phi(s_t) - \hat{R}_t\bigr)^2\right].
\end{equation}

The reward-side problem is now addressed: Actor-Critic provides per-timestep advantage estimates that remove pre-$t$ rewards and control variance through GAE.
Resolving it, however, reopens the trajectory side in a new form: stale on-policy data and unconstrained policy updates.

\section{PPO: Stability and Data Efficiency}
\label{sec:ppo_section}

Section~\ref{sec:actor_critic_section} produced the first complete per-timestep optimization form but left two trajectory-side problems open: stale on-policy data and unconstrained policy updates.
PPO resolves them through importance sampling, which enables data reuse across multiple gradient steps, and clipping, which constrains the policy shift to a safe trust region.

\subsection{Importance Sampling for Data Reuse}
\label{sec:is}

Importance sampling rewrites an on-policy expectation using samples from a different distribution.
Writing $\piold$ for the policy that collected the current data batch, the on-policy objective becomes
\begin{equation}
  L^{\mathrm{IS}}(\theta)
  = \E_{(s_t,a_t)\sim\piold}\!\left[r_t(\theta)\,A_t\right],
  \qquad
  r_t(\theta) = \frac{\pitheta(a_t|s_t)}{\piold(a_t|s_t)}.
\end{equation}
The ratio $r_t(\theta)$ corrects for the distribution mismatch, allowing the same rollout batch to be reused for multiple gradient steps.
As $\pitheta$ diverges from $\piold$, however, the weights become extreme and the estimator's variance grows without bound.
This variance growth is the update-instability problem, and it requires a mechanism to constrain how far the policy can move per update.

\subsection{PPO-Clip}
\label{sec:ppo_clip}

On-policy Actor-Critic variants such as A3C~\citep{mnih2016a3c} increase training throughput through asynchronous parallelism but do not resolve the data-efficiency problem: each rollout batch is still discarded after a single gradient step.
TRPO~\citep{schulman2015trpo} handles update instability through an explicit KL constraint, but this requires computing the Fisher information matrix, which is expensive for large networks.
PPO~\citep{schulman2017proximal} replaces the constraint with a clipping operation that removes the gradient incentive to push the probability ratio outside $[1-\epsilon, 1+\epsilon]$:
\begin{equation}
  \label{eq:ppo_clip}
  L^{\mathrm{CLIP}}(\theta)
  = \E_t\!\left[
      \min\!\left(
        r_t(\theta)\,A_t,\;
        \clip\!\left(r_t(\theta),1-\epsilon,1+\epsilon\right)A_t
      \right)
    \right].
\end{equation}
The $\min$ operator implements conservative pessimism.
When $A_t>0$, increasing $r_t$ beyond $1+\epsilon$ produces no additional gradient.
When $A_t<0$, pushing $r_t$ below $1-\epsilon$ yields no further gradient.
In both cases the policy can improve within the trust region but receives no signal to exceed it.

In practice, PPO preserves the actor-critic structure introduced in Section~\ref{sec:actor_critic}.
The actor replaces the vanilla policy-gradient objective with the clipped surrogate in Equation~\eqref{eq:ppo_clip}.
The critic is trained to minimize the squared error against the $\lambda$-return target $\hat{R}_t$ from Equation~\eqref{eq:lambda_return}, held fixed during the update of $\phi$:
\begin{equation}
  \label{eq:ppo_critic}
  L_{\mathrm{critic}}(\phi)
  =
  \E_t\!\left[
    \left(V_\phi(s_t)-\hat{R}_t\right)^2
  \right].
\end{equation}
Some implementations additionally clip the value-function update to prevent the critic from moving too far from its previous estimate.
Define the clipped value as
\begin{equation}
  V_t^{\mathrm{CLIP}} = \clip\!\bigl(V_\phi(s_t),\;V_t^{\mathrm{old}}-\epsilon_v,\;V_t^{\mathrm{old}}+\epsilon_v\bigr),
\end{equation}
where $\epsilon_v$ is a separate threshold from the policy-ratio $\epsilon$: the former bounds an absolute shift in value units, while the latter bounds a dimensionless probability ratio.
The clipped critic loss is then
\begin{equation}
  \label{eq:ppo_critic_clip}
  L_{\mathrm{critic}}^{\mathrm{clip}}(\phi)
  = \E_t\!\left[
      \max\!\left(
        \bigl(V_\phi(s_t) - \hat{R}_t\bigr)^2,\;
        \bigl(V_t^{\mathrm{CLIP}} - \hat{R}_t\bigr)^2
      \right)
    \right].
\end{equation}
The $\max$ retains whichever term has the larger squared error. The clipped term is selected only when $V_\phi(s_t)$ has moved outside the trust region $[V_t^{\mathrm{old}}-\epsilon_v, V_t^{\mathrm{old}}+\epsilon_v]$ on the same side as the target $\hat{R}_t$, so that a further update would push $V_\phi$ even farther from $V_t^{\mathrm{old}}$. In that case $V_t^{\mathrm{CLIP}}$ is constant with respect to $\phi$, so no gradient flows and movement in that direction is suppressed. If instead $V_\phi(s_t)$ has overshot past the target, the unclipped term has the larger error and stays active, so the gradient still pulls $V_\phi$ back toward $\hat{R}_t$. The clip therefore caps steps that drive the value away from its previous estimate. It does not block corrective ones.
This is a stabilization detail rather than the defining PPO-Clip mechanism, and many implementations omit it.

\subsection{Training Loop}
\label{sec:ppo_loop}

PPO proceeds in alternating rollout and update phases.
\begin{enumerate}
  \item Fix $\piold\leftarrow\pitheta$.  Roll out $\piold$ to collect a batch of trajectories.  Record states, actions, rewards, values, and action log-probabilities.
  \item Compute GAE advantages $A_t^{\mathrm{GAE}}$ using Equation~\eqref{eq:gae} and the $\lambda$-return targets $\hat{R}_t$ from Equation~\eqref{eq:lambda_return}.
  \item For $K$ epochs over the collected data, compute $r_t(\theta)=\pitheta(a_t|s_t)/\piold(a_t|s_t)$, evaluate $L^{\mathrm{CLIP}}$, update $\theta$, evaluate $L_{\mathrm{critic}}$ against $\hat{R}_t$, and update $\phi$.
  \item Return to step~1.
\end{enumerate}
The multi-epoch update in step~3 is possible because importance sampling decouples the gradient computation from the data-collecting policy.
This is what makes PPO data-efficient relative to vanilla Actor-Critic.

Importance sampling and clipping resolve both trajectory-side problems: data reuse is now possible across multiple gradient steps, and the clipping mechanism keeps the updated policy within a trust region where the advantage estimates remain valid.
Two things remain before PPO can be applied to LLMs.
First, the derivation so far assumes a generic MDP, and autoregressive token generation must be verified to satisfy the structural premises PPO requires.
Second, the reward model scores a complete response with a single scalar, but PPO requires per-token advantages. Bridging this mismatch calls for an explicit token-level reward construction.
\section{LLM as MDP: Why PPO Applies}
\label{sec:llm_mdp}

Both remaining questions concern how the generic MDP derivation connects to the specifics of autoregressive generation.
The trajectory-side question is whether token generation satisfies the Markov property and yields deterministic transitions, so that the policy gradient depends only on the policy and not on the environment dynamics, and whether action log-probabilities are differentiable through $\theta$.
The reward-side question is how a scalar response-level reward connects to the per-token credit assignment that PPO's GAE computation requires.
This section answers both and shows that the full derivation carries over to the LLM setting without modification.

\subsection{The MDP--LLM Correspondence}
\label{sec:llm_correspondence}

Given a prompt $x$, an autoregressive LLM generates a response $y=(y_1,\ldots,y_T)$ token by token.
Table~\ref{tab:mdp_llm} shows the exact correspondence between standard MDP variables and LLM generation.

\begin{table}[h]
\centering
\caption{MDP--LLM correspondence used throughout this survey.}
\label{tab:mdp_llm}
\setlength{\tabcolsep}{4pt}
\begin{tabular}{lll}
\toprule
\rowcolor{tabheadI}
MDP variable & LLM counterpart & Description \\
\midrule
State $s_t$ & $(x,\,y_{<t})$ & Prompt concatenated with generated prefix \\
Action $a_t$ & Token $y_t$ & Next token selected from the vocabulary \\
Policy $\pitheta(a_t|s_t)$ & $\pitheta(y_t|x,y_{<t})$ & Next-token distribution \\
Trajectory $\tau$ & $(x,\,y_{1:T})$ & Complete prompt-response pair \\
Transition $P(s_{t+1}|s_t,a_t)$ & $s_{t+1}=(x,y_{\le t})$ & Deterministic string concatenation \\
Reward $R(\tau)$ & $R(x,y)$ & Reward model or verifier score \\
\bottomrule
\end{tabular}
\end{table}

Token generation can be modeled as an episodic, finite-horizon task. Because the episode always terminates at the end-of-sequence token, the infinite-horizon convergence requirement that enforces $\gamma<1$ does not apply, and we set $\gamma=1$ so that future rewards are not discounted by token position. 
In practice, the state space $(x, y_{<t})$ is extremely large, so all expectations and gradients are estimated via sampling rather than exact enumeration.
This simplification is standard in LLM policy optimization~\citep{ziegler2019fine,ouyang2022training}.
From this section onwards all LLM-specific formulations implicitly assume $\gamma=1$ for the return, so the reward-to-go reduces to $G_t=\sum_{k\ge t} r_k$ and no token position is discounted. The GAE expressions from Part~I carry over unchanged. We note that the $\gamma$ inside GAE need not equal the return discount. Several RLHF implementations keep a $\gamma$ slightly below one in the GAE recursion as a variance-control knob on long sequences, while still leaving the terminal reward undiscounted. The role of $\lambda$ as the bias--variance control is the same in either case.

Two structural features make the MDP formulation exact at the level of autoregressive token generation.
First, the state transition is deterministic: given \(s_t=(x,y_{<t})\) and \(a_t=y_t\), the next state is \(s_{t+1}=(x,y_{\le t})\).
Since this transition kernel is independent of \(\theta\), it contributes no score term to the policy-gradient estimator, exactly as in Section~\ref{sec:pg}.
Second, the log-probability factorizes as \[
\log \pi_\theta(y\mid x)
=
\sum_t \log \pi_\theta(y_t\mid x,y_{<t}),
\]
which is computable by a standard forward pass and differentiable with respect to \(\theta\).
These properties ensure that autoregressive generation admits an exact policy-gradient formulation with token-level actions and prefix states.

\subsection{The RLHF Pipeline}
\label{sec:rlhf}

RLHF~\citep{stiennon2020learning,ouyang2022training} typically proceeds in three stages.
\begin{enumerate}
  \item \textbf{Supervised fine-tuning.} A pretrained LLM is fine-tuned on human demonstrations to obtain the initial policy, and a frozen copy of this model or the pretrained base model serves as the reference model $\piref$.
  \item \textbf{Reward model training.} Human annotators compare pairs of responses to the same prompt $x$ and identify the preferred response $y_w$ over the rejected response $y_l$.
        A scalar reward model $R_\psi(x,y)$~\citep{bai2022training,christiano2017deep} is trained to assign higher scores to preferred responses under the Bradley--Terry model~\citep{bradley1952rank}:
        \begin{equation}
          \label{eq:bt_model}
          P(y_w \succ y_l \mid x)
          = \sigma\!\bigl(R_\psi(x,y_w) - R_\psi(x,y_l)\bigr),
        \end{equation}
        where $\sigma$ denotes the sigmoid function, and the training loss is the negative log-likelihood of the observed preferences.
  \item \textbf{RL fine-tuning.} The actor is optimized to maximize the reward model score while being regularized by a KL penalty against $\piref$:
        \begin{equation}
          \label{eq:rlhf_obj}
          \max_\theta\;
          \E_{x,\,y\sim\pitheta}\!\left[
            R_\psi(x,y)
            - \beta\,\KL(\pitheta(\cdot|x)\|\piref(\cdot|x))
          \right].
        \end{equation}
\end{enumerate}
In practice, the sequence-level KL term is implemented through sampled per-token log-ratio penalties, as made explicit in Section~\ref{sec:token_reward}.

\subsection{Four-Component Architecture}
\label{sec:four_components}

The four model roles in LLM RLHF-PPO are summarized in Table~\ref{tab:components}.
The critic and the reward model are distinct. 
The critic estimates future value from partial generations, while the reward model evaluates complete responses.

\begin{table}[h]
\centering
\caption{Four model components in LLM RLHF-PPO and their roles.}
\label{tab:components}
\begin{tabular}{llc}
\toprule
\rowcolor{tabheadI}
Component & Role & Updated during RL? \\
\midrule
Actor $\pitheta$     & Policy that generates tokens         & Yes \\
Critic $V_\phi$ & Estimates per-token state value $V^\pi(s_t)$ & Yes \\
Reward model $R_\psi$& Scores complete responses            & No  \\
Reference $\piref$   & Provides KL regularization          & No  \\
\bottomrule
\end{tabular}
\end{table}

\subsection{Token-Level Reward from Sequence-Level Score}
\label{sec:token_reward}

The reward model outputs a single scalar for the complete response, but PPO must update each token's probability.
To bridge this mismatch, we define a token-level reward sequence: the final token receives the full sequence reward from the reward model, while earlier tokens receive only per-token KL penalties.
Generalized Advantage Estimation (GAE) combines this reward sequence with the critic's value estimates to assign credit to earlier token decisions through temporal-difference residuals.
For long sequences or sparse rewards, the variance of these advantages can still be high, which motivates alternative baseline strategies in Part~II.
Formally,
\begin{equation}
  \label{eq:token_reward}
  r_t =
  \begin{cases}
    R_\psi(x,y)
    - \beta\log\dfrac{\pitheta(y_t|x,y_{<t})}{\piref(y_t|x,y_{<t})}
    & t = T, \\[6pt]
    -\beta\log\dfrac{\pitheta(y_t|x,y_{<t})}{\piref(y_t|x,y_{<t})}
    & t < T.
  \end{cases}
\end{equation}
Here the KL penalty is written with $\pitheta$ to denote the policy that generated the response. In practice the entire token-level reward sequence, including the KL term, is computed once at rollout time and then held fixed. GAE turns this sequence into token-level advantages. These advantages are reused across the $K$ inner update epochs rather than recomputed at each step~\citep{ouyang2022training}. At collection time $\piold\leftarrow\pitheta$, so the policy appearing here coincides with $\piold$. This matches the sequence-level KL objective in~\eqref{eq:rlhf_obj}.
GAE applied to this sequence with $V_\phi$ produces token-level advantages $A_t^{\mathrm{GAE}}$, and the PPO-Clip objective~\eqref{eq:ppo_clip} is applied per token with
\[
r_t(\theta)=
\dfrac{\pitheta(y_t|x,y_{<t})}{\piold(y_t|x,y_{<t})}.
\]

\subsection*{Part~I Summary}
\label{sec:part1_summary}

Part~I began from the intractable gradient of $J(\theta)$ and resolved problems on each side in sequence.
On the trajectory side, the log-derivative trick converted the intractable gradient into a sampling-friendly expectation. The environment dynamics terms vanish because they do not depend on $\theta$, and in the LLM setting this holds exactly because transitions are deterministic.
Importance sampling and PPO-Clip then resolved the data-reuse and large-update instability problems that arise when attempting to reuse on-policy Actor-Critic data across multiple gradient steps.
On the reward side, reward-to-go, advantage estimation, and GAE replaced the noisy trajectory return with causally correct, variance-reduced per-timestep advantages.
The MDP--LLM correspondence confirmed that token generation satisfies every structural premise the derivation requires, so the full derivation carries over to the LLM setting intact.
A remaining challenge is that the per-token critic $V_\phi$ must match the actor in capacity, roughly doubling memory and compute, and becomes numerically unstable under the sparse or binary rewards typical of LLM training.
In such cases, value targets are noisy or poorly calibrated, producing high-variance or biased advantage estimates and potentially destabilizing PPO updates.

Part~II takes this per-step form as its starting point and addresses the critic challenge directly: GRPO performs a reward-side substitution, replacing the per-token critic-estimated advantage with a group-relative scalar computed from sampled rewards.
Part~III then extends the framework along both axes: Agentic RL generalizes the trajectory side to multi-turn environment interaction, while the GRPO-OPD hybrid incorporates a dense per-token teacher signal into the reward objective.

\part*{Part~II\quad LLM Policy Optimization in Practice}
\addcontentsline{toc}{part}{Part II: LLM Policy Optimization in Practice}

\section{GRPO and the Single-Turn Landscape}
\label{sec:grpo_line}

Part~I established the per-step LLM optimization objective: a per-token clipped importance-ratio objective, combined with per-token critic-estimated advantages and a KL penalty on the objective.
Although this formulation is theoretically sound, it is computationally expensive.
The critic required for per-token advantage estimation must be comparable in scale to the actor.
For long chain-of-thought tasks with sparse or binary rewards, this requirement is both computationally prohibitive and numerically unstable.

Throughout this survey, \emph{reward side} refers to the full pipeline from reward signal to advantage estimate: how rewards are obtained, normalized, shaped, and converted into the scalar that weights each policy-gradient step.
The main founding contribution of GRPO is a reward-side innovation: the computational cost of the per-token critic arises from advantage estimation, not from trajectory sampling.
Therefore, eliminating this cost requires only a reward-side substitution while leaving the trajectory side unchanged.
This observation provides the starting point for Part~II.
Subsequent work after GRPO reveals limitations on both sides, and later methods address these failures either independently or jointly, while remaining within the single-turn token-generation setting.
Part~III then examines two structural extensions: Agentic RL, where the trajectory side expands from single-turn generation to multi-turn environment interaction, and the GRPO-OPD hybrid, where the reward side incorporates a dense per-token teacher signal as a component of the reward objective while retaining $J(\theta)$.

\begin{figure}[t]
\centering
\resizebox{\linewidth}{!}{%
\begin{tikzpicture}[
  x=1cm, y=1cm,
  ROOT/.style={
    draw=blue!55, fill=blue!8, rounded corners=4pt, thick,
    font=\small\bfseries, align=center,
    minimum width=3.7cm, minimum height=1.0cm, inner sep=5pt
  },
  SIDE/.style={
    draw=cTraj!60, fill=cTraj!6, rounded corners=4pt, thick,
    font=\small\bfseries, align=center,
    minimum width=3.1cm, minimum height=0.9cm, inner sep=5pt
  },
  PROB/.style={
    draw=gray!40, fill=gray!4, rounded corners=3pt,
    font=\footnotesize, align=left,
    text width=8.35cm, inner sep=5pt
  },
  arr/.style={-{Stealth[length=5pt,width=4pt]}, thick, gray!60},
  barr/.style={-{Stealth[length=5pt,width=4pt]}, thick, blue!50},
  bline/.style={thick, blue!50},
]

\def\probGap{0.18cm}   

\def\rootX{1.60}       
\def\stubX{4.55}       
\def\sideX{6.70}       
\def\leafX{8.95}       


\node[PROB, draw=cTraj!45, fill=cTraj!4, anchor=west] (traj_leaf) at (\leafX, 2.90) {%
  \textcolor{cTraj!85!black}{\textbf{Collection:}}
    GRESO~\citep{zheng2026greso},
    Prompt Replay~\citep{prompt_replay2025},
    PODS~\citep{pods2025},
    Adap.\ rollout~\citep{adaptive_rollout2025,arlr2025}\\[2pt]
  \textcolor{cTraj!85!black}{\textbf{Compression \& reuse:}}
    ThinkPrune~\citep{thinkprune2025},
    DAST~\citep{dast2025},
    RePO~\citep{repo2025},
    ExGRPO~\citep{exgrpo2025}\\[2pt]
  \textcolor{cTraj!85!black}{\textbf{Diversity:}}
    TreeRPO~\citep{yang2025treerpo},
    DIVER~\citep{diver2025},
    DRA-GRPO~\citep{dragrpo2025},
    DSDR~\citep{dsdr2026},
    R1-zero-Div~\citep{dmpo2025},
    GAPO~\citep{anschel2025gapo},
    Pro-GRPO~\citep{progrpo2025},
    DaGRPO~\citep{dagrpo2025}\\[2pt]
  \textcolor{cTraj!85!black}{\textbf{Repair:}}
    S2R~\citep{s2r2025},
    SCoRe~\citep{score2025},
    PAG~\citep{pag2025}\\[2pt]
  \textcolor{cTraj!85!black}{\textbf{Clip \& ratio granularity:}}
    DAPO~\citep{yu2026dapo},
    SAPO~\citep{gao2025sapo},
    GSPO~\citep{zheng2025gspo}
};

\node[PROB, draw=cRew!40, fill=cRew!4, anchor=north west] (rew_leaf)
  at ([yshift=-\probGap]traj_leaf.south west) {%
  \textcolor{cRew}{\textbf{Density:}}
    GRPO-VPS~\citep{grpovps2026},
    SPO~\citep{guo2025spo}\\[2pt]
  \textcolor{cRew}{\textbf{Source:}}
    Open-Reasoner-Zero~\citep{openreasoner2025},
    R1-V~\citep{chenr1},
    One-shot RLVR~\citep{oneshot_rlvr2025}\\[2pt]
  \textcolor{cRew}{\textbf{Shaping:}}
    LASER~\citep{laser2025},
    S-GRPO~\citep{sgrpo2025}\\[2pt]
  \textcolor{cRew}{\textbf{Multi-objective:}}
    QwQ~\citep{qwq2025},
    Kimi~k1.5~\citep{kimi2025k15},
    GLM-4.5V~\citep{glmv2025}\\[2pt]
  \textcolor{cRew}{\textbf{Stat.\ bias corr.:}}
    Dr.GRPO~\citep{liu2025understanding},
    Lopti~\citep{yang2025not},
    NGRPO~\citep{nan2025ngrpo},
    AGPO~\citep{li2025agpo},
    CPPO~\citep{cppo2025},
    RiskPO~\citep{ren2025riskpo}\\[2pt]
  \textcolor{cRew}{\textbf{Credit assign.:}}
    TreeRPO~\citep{yang2025treerpo},
    SCRL~\citep{scrl2026},
    SPO~\citep{guo2025spo},
    HighEnt~\citep{wang2025entropy_token},
    PRIME~\citep{cui2025process},
    GRPO-$\lambda$~\citep{grpo_lambda2025},
    VAPO~\citep{vapo2025},
    GRPO-VPS~\citep{grpovps2026}
};

\node[PROB, draw=cBoth!35, fill=cBoth!4, anchor=north west] (both_leaf)
  at ([yshift=-\probGap]rew_leaf.south west) {%
  \textcolor{cBoth}{\textbf{Clip--variance coupling:}}
    DAPO~\citep{yu2026dapo},
    DCPO~\citep{yang2025dcpo}\\[2pt]
  \textcolor{cBoth}{\textbf{Granularity coupling:}}
    HT-GRPO~\citep{htgrpo2025},
    GTPO~\citep{gtpo2025}
};


\node[SIDE] (traj) at (\sideX,0 |- traj_leaf)
  {Trajectory-Side\\Extensions};

\node[SIDE, draw=cRew!55, fill=cRew!5] (rew) at (\sideX,0 |- rew_leaf)
  {Reward-Side\\Extensions};

\node[SIDE, draw=cBoth!50, fill=cBoth!5] (both) at (\sideX,0 |- both_leaf)
  {Both-Side\\Extensions};


\node[ROOT] (grpo) at (\rootX,0 |- rew)
  {GRPO~\citep{shao2024deepseekmath}\\\scriptsize\itshape reward-side: critic-free baseline};


\coordinate (stub)      at (\stubX,0 |- grpo);
\coordinate (stub_traj) at (\stubX,0 |- traj);
\coordinate (stub_rew)  at (\stubX,0 |- rew);
\coordinate (stub_both) at (\stubX,0 |- both);

\draw[bline] (grpo.east) -- (stub);

\draw[bline] (stub_traj) -- (stub_both);

\draw[barr] (stub_traj) -- (traj.west);
\draw[barr] (stub_rew)  -- (rew.west);
\draw[barr] (stub_both) -- (both.west);


\draw[arr, cTraj!60] (traj.east) -- (traj_leaf.west);
\draw[arr, cRew!55]  (rew.east)  -- (rew_leaf.west);
\draw[arr, cBoth!50] (both.east) -- (both_leaf.west);

\end{tikzpicture}%
}
\caption{%
  Structure of the GRPO line (\S\ref{sec:grpo_line}).
  GRPO~\citep{shao2024deepseekmath} originates as a pure reward-side substitution:
  the group-relative baseline $\hat{A}_i=(R_i-\mu_G)/\sigma_G$ replaces the trained critic
  while the trajectory probability side (importance ratio, clip, KL penalty) is unchanged.
  Post-GRPO work identifies trajectory-side failures (collection, compression and reuse,
  diversity, repair, and clip--ratio granularity mismatch), reward-side failures
  (density, source, shaping, multi-objective aggregation, statistical bias correction,
  and credit assignment), and compound instabilities that require joint intervention
  on both sides.
}
\label{fig:grpo_line_overview}
\end{figure}
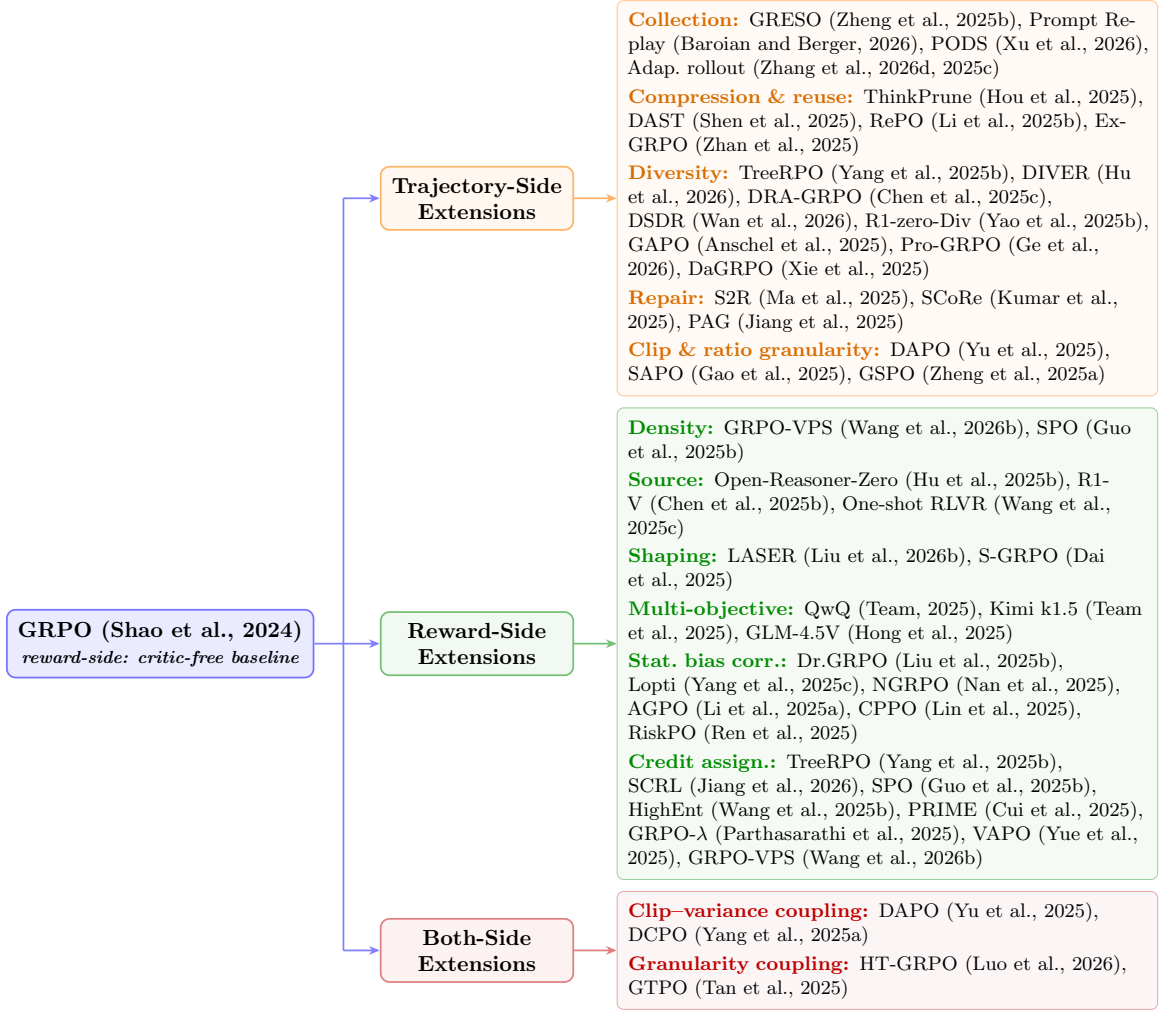

\subsection{GRPO: Eliminating the Critic}
\label{sec:grpo}

\paragraph{Core idea.}
Group Relative Policy Optimization (GRPO)~\citep{shao2024deepseekmath} replaces the value model with group-relative reward normalization.
For a prompt $x$, $G$ responses $y_1,\ldots,y_G$ are sampled from $\piold$ and evaluated by a verifier or reward model, yielding $R_i=R(x,y_i)$.
The group-relative advantage is defined as
\begin{equation}
  \label{eq:grpo_adv}
  \hat{A}_i = \frac{R_i - \mu_G}{\sigma_G},
  \qquad
  \mu_G = \tfrac{1}{G}\textstyle\sum_{i}R_i,
  \quad
  \sigma_G = \!\sqrt{\tfrac{1}{G}\textstyle\sum_{i}(R_i-\mu_G)^2}.
\end{equation}
This scalar advantage is then broadcast uniformly to every token in $y_i$.
The objective is
\begin{equation}
  \label{eq:grpo_obj}
  L^{\mathrm{GRPO}}(\theta)
  = \E\!\left[
      \frac{1}{G}\sum_{i=1}^{G}
      \frac{1}{|y_i|}\sum_{t=1}^{|y_i|}
      \min\!\bigl(r_{i,t}\hat{A}_i,\;
                  \clip(r_{i,t},1-\epsilon,1+\epsilon)\hat{A}_i\bigr)
    \right]
  - \beta\,\KL(\pitheta\|\piref),
\end{equation}
where $r_{i,t}=\pitheta(y_{i,t}|x,y_{i,<t})/\piold(y_{i,t}|x,y_{i,<t})$ denotes the per-token importance ratio, which is identical to the RLHF-PPO ratio in Equation~\eqref{eq:ppo_clip}.
A comparison between GRPO and RLHF-PPO makes the modification precise: the trajectory side remains unchanged, with the same importance ratio, the same symmetric clipping rule, and the same KL regularization toward $\piref$, whereas the reward side is restructured from a critic-estimated per-token advantage $A_t^{\mathrm{GAE}}$ into a group-relative scalar advantage $\hat{A}_i$.
The two formulations place this KL term differently. RLHF-PPO folds it into the per-token reward sequence in Equation~\eqref{eq:token_reward}, whereas GRPO writes it as a separate penalty on the objective in Equation~\eqref{eq:grpo_obj}. The regularization role is the same, only the point of insertion differs.
GRPO is therefore a purely reward-side substitution.

GRPO belongs to a broader family of critic-free methods that replace the value network with a Monte Carlo baseline computed from sampled responses.
RLOO~\citep{ahmadian2024back} builds the baseline for each response by averaging the returns of the other responses to the same prompt, a leave-one-out estimate.
ReMax~\citep{li2024remax} instead subtracts the reward of a single greedy response.
REINFORCE++~\citep{hu2025reinforce} keeps a global baseline and adds normalization and variance-reduction techniques to the basic REINFORCE estimator.
What sets GRPO apart is the division of the centered reward by the within-group standard deviation $\sigma_G$. This extra step strips out the reward scale, and it is the origin of the statistical biases examined in Section~\ref{sec:stat_bias}.

DeepSeek-R1~\citep{guo2025deepseek} demonstrated that this substitution, when combined with verifiable rewards, can produce strong mathematical reasoning.
Subsequent analyses~\citep{rlvr2025incentivize,qwen25math} clarified the implicit incentive structure, and open reasoning models such as QwQ-32B~\citep{qwq2025} adopted GRPO-style RLVR at scale.

Recent surveys have documented the rapid proliferation of RLVR methods after DeepSeek-R1.
A technical comparison of PPO, GRPO, and their variants is provided by \citet{srivastava2025technical,wang2024reinforcement}.
The broader landscape of agentic RL, which extends these methods to multi-turn environments, is surveyed by \citet{zhang2025landscape}.
Practical training recipes and analyses of hyperparameter sensitivity are reported by \citet{openreasoner2025}.

\paragraph{The failure structure.}
The deficiencies of GRPO align closely with the two algorithmic sides.
On the trajectory side, failures arise in two places.
Some failures concern the trajectories themselves before any update: uniform prompt sampling wastes compute on uninformative groups, long responses introduce length-variance artifacts, group diversity decreases as the policy converges, and failed rollouts provide no gradient.
Other failures concern how the update rule of GRPO consumes these trajectories.
Hard symmetric clipping, combined with group-relative normalization, can drive entropy collapse by constraining low-probability exploratory tokens more aggressively than high-probability ones as the group distribution narrows. This coupling is treated as a compound failure in Section~\ref{sec:both_variants}.
Separately, pairing the token-level importance ratio $r_{i,t}$ with the sequence-level advantage $\hat{A}_i$ creates a structural granularity mismatch.

On the reward side, limitations arise at two levels: the reward signal $R_i$ itself, and the advantage $\hat{A}_i$ that converts it into a per-token weight.
At the reward-signal level, the sparse scalar reward $R_i$ can be too coarse to identify which steps or tokens determine the outcome. Without additional design, it is also insensitive to output length, format, diversity, and competing objectives.
At the advantage level, normalization by $\sigma_G$ and $|y_i|$ introduces statistical bias.
When all responses in a group receive the same reward, the centered numerator $R_i-\mu_G$ and the denominator $\sigma_G$ both vanish, so the normalized advantage is formally $0/0$. 
Implementations resolve this by adding a small constant to $\sigma_G$ or by skipping the group, which sets the advantage to zero and leaves the gradient empty. Section~\ref{sec:stat_bias} treats this degeneracy in detail.
Beyond normalization, the scalar advantage $\hat{A}_i$ is assigned identically to every token, treating all positions as equally responsible for the outcome: this is the credit assignment problem, and it motivates a separate class of methods that replace the uniform broadcast with position-sensitive signals.

The subsections below group these failures by side, beginning with the trajectory side, then the reward side, and finally methods that address failures on both sides jointly.
All methods discussed in this section remain within the single-turn token-generation setting.
Part~III then considers two structural departures from this endpoint: Section~\ref{sec:agentic} expands the trajectory side by replacing single-turn generation with multi-turn environment interaction, and Section~\ref{sec:opd_line} extends the reward side by incorporating a dense per-token teacher signal as a component of the reward objective while retaining $J(\theta)$.

\subsection{Trajectory-Side Variants}
\label{sec:traj_variants}

From the trajectory perspective, the central question can be decomposed into two sub-questions that jointly determine which learning signals contribute to the gradient.
First, \emph{which trajectories should be used}. 
Standard uniform rollout collection can spend rollout budget on uninformative prompts, reduce group diversity, or produce trajectories whose processing cost is prohibitive without compression. 
Collection, compression, diversity, and repair methods address this sub-question by determining which trajectories are included in the update and in what form.
Second, \emph{how much each trajectory should be optimized}. 
The clipping rule and ratio granularity define the trust region. 
They control the magnitude of the update induced by an individual trajectory and the level of aggregation at which the correction is applied.

\paragraph{Working within the GRPO update rule.}
These methods keep the importance ratio and the clipping rule fixed, and they address the first sub-question above, namely which trajectories should be used.
The design space can be organized along four dimensions.
Trajectory collection determines which prompts and rollouts are sampled. 
Trajectory compression and reuse constrain trajectories and amortize sampling cost. 
Trajectory diversity preserves contrastive signals within the group. 
Trajectory repair recovers useful signals from failed rollouts.

\subsubsection*{\textbf{Trajectory Collection}}
\phantomsection\label{sec:traj_collection}

\textbf{Problem.} Standard GRPO samples $G$ responses uniformly, but many prompts yield zero-variance groups and therefore no useful advantage signal~\citep{zheng2026greso,yu2026dapo}.
Better collection strategies concentrate sampling on prompts with high group variance and strong gradient signal.

GRESO~\citep{zheng2026greso} predicts likely uninformative prompts from historical reward dynamics and filters them before rollout.
Prompt Replay~\citep{prompt_replay2025} maintains a buffer of high-signal prompts and mixes them with fresh samples in each batch.
New responses are always sampled from the current policy, so the importance ratio remains valid regardless of whether the prompt came from the buffer or from fresh sampling.
PODS~\citep{pods2025} can also be viewed as a rollout-selection strategy. It generates a larger set of responses per prompt and updates only on a down-sampled subset chosen to maximize reward variance, thereby preserving contrastive signal.
Adaptive rollout-allocation methods~\citep{adaptive_rollout2025,arlr2025} allocate more samples to prompts with uncertain or non-saturated rewards and fewer samples to already-converged prompts.
Together, these methods redirect rollout compute from zero-signal groups toward higher-gradient updates without fundamentally changing the overall GRPO-style update.
In summary, collection methods solve the informative-group problem by filtering or prioritizing which prompts or rollouts enter the update.

\subsubsection*{\textbf{Trajectory Compression and Reuse}}
\phantomsection\label{sec:traj_compression}

\textbf{Problem.} Rolling out $G$ fresh on-policy responses for each prompt at each update step and processing long reasoning chains constitute the dominant training costs in GRPO-style training~\citep{thinkprune2025,repo2025}.
A secondary issue arises from length variance. When response lengths vary substantially within a group, the group-normalized advantage becomes sensitive to length, which introduces noise that is unrelated to correctness.
Reusing past response trajectories amortizes sampling cost but introduces off-policy drift, which must be controlled.

ThinkPrune~\citep{thinkprune2025} shortens reasoning by training under explicit token budgets and progressively tightening these budgets.
DAST~\citep{dast2025} uses difficulty-adaptive reasoning budgets, which encourage shorter reasoning for easy prompts while preserving longer reasoning for hard prompts. The difficulty signal gates trajectory length rather than modifying the reward, so DAST remains a trajectory-side compression method.
RePO~\citep{repo2025} extends GRPO by maintaining a replay buffer of previously sampled outputs and retrieving diverse off-policy samples for each prompt. It combines on-policy and off-policy terms to improve data efficiency.
ExGRPO~\citep{exgrpo2025} maintains a prioritized experience buffer ranked by correctness and response entropy, and selectively replays trajectories whose signals would otherwise be lost after a single update step.
In summary, both families reduce training cost. Compression shortens trajectories to reduce per-step computation, whereas reuse amortizes the cost of fresh rollouts across multiple updates. Both families must satisfy a correctness constraint. Stronger compression may discard informative tokens, and reuse must keep importance-ratio drift within bounds to preserve a valid policy-gradient estimate.

\subsubsection*{\textbf{Trajectory Diversity}}
\phantomsection\label{sec:traj_diversity}

\textbf{Problem.} When the policy collapses to a narrow set of solution strategies, the sampled group may lose semantic and reward diversity, yielding weak or redundant group-relative learning signals even when some responses are correct~\citep{anschel2025gapo,dragrpo2025,progrpo2025}.
Without explicit diversity pressure, the rollout distribution can become degenerate, and the group may no longer provide useful contrastive signals, regardless of reward quality.

TreeRPO~\citep{yang2025treerpo} uses tree-structured sampling rather than independent linear rollouts.
From each intermediate node, it samples multiple candidate continuations and forms step-level groups from sibling branches.
The branching structure increases trajectory-level exploration and exposes structurally different reasoning prefixes during training.
Because sibling branches also enable step-level expected reward estimation, TreeRPO doubles as a credit assignment mechanism and is revisited in Section~\ref{sec:credit_assign}.
DIVER~\citep{diver2025} introduces global sequence-level diversity as an intrinsic reward and incorporates it through potential-based reward shaping, preserving optimal-policy invariance while encouraging deeper exploration in semantically structured reasoning spaces.
DRA-GRPO~\citep{dragrpo2025} acts after trajectories have been sampled. 
It estimates semantic density within the sampled group and uses a diversity-aware reward adjustment to down-weight redundant completions and amplify rewards for structurally distinct reasoning paths.

DSDR~\citep{dsdr2026} also operates on the collected trajectory set by adding dual-scale diversity regularization to the training objective. At the global level, it encourages distinct solution modes among correct reasoning trajectories. At the local level, it applies length-invariant token-level entropy regularization restricted to correct trajectories, preventing collapse within each mode while preserving correctness.
R1-zero-Div~\citep{dmpo2025} adds a token-level diversity objective to the GRPO-style policy objective and applies it selectively to positive samples, motivated by the observed positive correlation between solution diversity and reasoning potential.
GAPO~\citep{anschel2025gapo} extends GRPO by computing rewards over group-level properties such as diversity and coverage. With frequency-aware group rewards, it penalizes over-represented completions and encourages more uniform coverage of valid responses.

Pro-GRPO~\citep{progrpo2025}, developed for GRPO-style training in generative models, addresses reward clustering by expanding the initial sampling group and pruning reward-clustered trajectories during sampling using latent feature proxies. This dynamic expand-and-prune strategy retains a higher-variance survivor set while avoiding the full cost of generating trajectories that would later be discarded.
DaGRPO~\citep{dagrpo2025} identifies low-distinctiveness samples or response pairs in the collected group as a source of gradient conflict and dynamically masks them from the update. It also injects high-quality off-policy anchors to recover training signals on difficult prompts.
Together, these methods address diversity failures by shaping trajectory generation, adding diversity-aware rewards or regularizers to the training objective, or filtering and reweighting the collected trajectory set before the update.
They help maintain useful contrastive signals throughout training.

\subsubsection*{\textbf{Trajectory Repair}}
\phantomsection\label{sec:traj_repair}

\textbf{Problem.} Outcome-only RLVR treats a response primarily through its final correctness, so a failed first attempt provides little information about where the reasoning went wrong or how the policy could recover.
In group-relative objectives, this inefficiency is amplified when all sampled trajectories for a prompt fail, since the group provides little or no reward contrast and the resulting advantage signal becomes uninformative.

Repair methods address this by expanding a failed single-shot response into a multi-turn attempt--verification--revision process, exposing correction behavior that is unavailable in single-shot outcome-only rollouts.
S2R~\citep{s2r2025} learns self-directed trial-and-error by combining self-verification and self-correction with outcome- and process-level reinforcement learning.
SCoRe~\citep{score2025} trains self-correction through multi-turn online RL on the model's own correction traces, using regularization and reward shaping to encourage genuine improvement over the first attempt.
PAG~\citep{pag2025} frames the same model as both policy and generative verifier, revising only when self-verification detects an error and thereby avoiding unnecessary second attempts.
From a GRPO-style training perspective, repair methods can convert otherwise terminal failures into structured verification--revision traces, potentially recovering learning signal from rollouts that would provide little contrast under outcome-only feedback.

\paragraph{Modifying the GRPO update rule.}
These methods accept trajectories as given and instead modify how GRPO processes them during the update step.
The key dimension of variation is the clipping rule, specifically how the trust region is enforced and at what granularity the importance ratio is computed.

\subsubsection*{\textbf{Clipping Rule and Ratio Granularity Modification}}
\phantomsection\label{sec:clip_rule}

\textbf{Problem.} The standard symmetric clipping rule $\clip(r_{i,t}, 1-\epsilon, 1+\epsilon)$ applies identical thresholds to upward and downward deviations of the importance ratio.
This symmetric bound can disproportionately restrict the probability increase of low-probability exploratory tokens with positive advantage, while high-probability exploitation tokens remain relatively easy to reinforce. This creates entropy-reduction pressure and can lead to premature determinism or mode collapse at scale~\citep{yu2026dapo}.
A separate problem arises from granularity: standard GRPO computes a per-token importance ratio, clips each token independently, and then multiplies the clipped ratio by the sequence-level advantage $\hat{A}_i$.
For a sequence-level reward, the policy gradient theorem requires a sequence-level probability correction. Applying token-level ratios to a sequence-level advantage therefore creates a systematic mismatch~\citep{zheng2025gspo}.

DAPO~\citep{yu2026dapo} decouples the lower and upper clipping thresholds by setting $\epsilon_{\mathrm{low}} < \epsilon_{\mathrm{high}}$.
The asymmetric upper bound gives low-probability but positively advantaged tokens more room to increase, mitigating the tendency of the policy to become prematurely deterministic.
DAPO also drops the KL penalty term entirely. During long reasoning training the policy is expected to move far from the initial model, so a fixed reference anchor restricts the very exploration the method tries to encourage.
Because DAPO further combines dynamic sampling, token-level policy-gradient loss, and overlong reward shaping, it addresses reward-side variance coupling as well and is revisited as a compound instability in Section~\ref{sec:both_variants}.
SAPO~\citep{gao2025sapo} replaces the hard clipping rule with a continuous soft gate. The gradient is attenuated smoothly as $r_{i,t}$ moves away from $1$, preserving useful signal from moderately off-policy tokens that hard clipping would otherwise discard.
GSPO~\citep{zheng2025gspo} resolves the granularity mismatch by clipping the sequence-level ratio directly:
\begin{equation}
  \label{eq:seq_ratio}
  r^{\mathrm{seq}}_i(\theta)
    =
    \left(
    \frac{\pitheta(y_i|x)}
         {\piold(y_i|x)}
    \right)^{1/|y_i|}
    =
    \exp\left(
    \frac{1}{|y_i|}
    \sum_{t=1}^{|y_i|}
    \log
    \frac{\pitheta(y_{i,t}|x,y_{i,<t})}
         {\piold(y_{i,t}|x,y_{i,<t})}
    \right).
\end{equation}
Clipping $r^{\mathrm{seq}}_i$ rather than individual $r_{i,t}$ aligns the level of probability correction with the level of the advantage signal. At coarser granularities such as step, sentence, or turn, the ratio is computed as a partial product over the tokens within that unit.
Clipping and granularity methods operate entirely within the update step and leave rollout trajectories unchanged. Their corrections are therefore orthogonal to the collection, compression, diversity, and repair interventions above and can in principle be combined with any of them.
In summary, asymmetric clipping and soft gating address the entropy-collapse pressure from the trust-region rule, while sequence-level ratio clipping resolves the structural mismatch between the granularity of the importance ratio and the granularity of the advantage.

\begin{table}[t]
\centering
\caption{Trajectory-side post-GRPO methods, grouped by whether they work within or modify the GRPO update rule.}
\label{tab:traj_variants}
\scriptsize
\setlength{\tabcolsep}{4pt}
\renewcommand{\arraystretch}{1.05}
\begin{tabularx}{\textwidth}{>{\raggedright\arraybackslash}p{3.0cm}>{\raggedright\arraybackslash}p{2.6cm}X}
\toprule
\rowcolor{tabheadTraj}
Method & Dimension & Key mechanism \\
\midrule
\rowcolor{tabbandTraj}
\multicolumn{3}{l}{\textit{Within GRPO update rule}} \\
\rowcolor{cTraj!8}
GRESO~\citep{zheng2026greso}               & Collection            & Informativeness prediction filters uninformative prompts. \\
\rowcolor{cTraj!2}
Prompt Replay~\citep{prompt_replay2025}    & Collection            & High-signal prompt buffer with fresh on-policy responses. \\
\rowcolor{cTraj!8}
PODS~\citep{pods2025}                      & Collection            & Over-generate then down-sample to a subset maximizing reward variance. \\
\rowcolor{cTraj!2}
Adap.\ rollout~\citep{adaptive_rollout2025,arlr2025} & Collection  & Difficulty-proportional group size allocation. \\
\addlinespace[2pt]
\rowcolor{cTraj!8}
ThinkPrune~\citep{thinkprune2025}          & Compression \& reuse  & Truncate chains beyond learned length threshold. \\
\rowcolor{cTraj!2}
DAST~\citep{dast2025}                      & Compression \& reuse  & Difficulty-adaptive length budget, short for easy and long for hard. \\
\rowcolor{cTraj!8}
RePO~\citep{repo2025}                      & Compression \& reuse  & Verified trajectory replay buffer under IS correction. \\
\rowcolor{cTraj!2}
ExGRPO~\citep{exgrpo2025}                  & Compression \& reuse  & Prioritised experience buffer by correctness and entropy. \\
\addlinespace[2pt]
\rowcolor{cTraj!8}
TreeRPO~\citep{yang2025treerpo}            & Diversity             & Branch rollouts from each intermediate step. Sibling branches also enable step-level reward estimation (also in Table~\ref{tab:reward_variants}). \\
\rowcolor{cTraj!2}
DIVER~\citep{diver2025}                    & Diversity             & Global diversity as intrinsic reward via potential-based shaping. \\
\rowcolor{cTraj!8}
DRA-GRPO~\citep{dragrpo2025}               & Diversity             & Diversity-aware reward adjustment down-weights redundant completions. \\
\rowcolor{cTraj!2}
DSDR~\citep{dsdr2026}                      & Diversity             & Dual-scale diversity regularization: global mode diversity + local token entropy. \\
\rowcolor{cTraj!8}
R1-zero-Div~\citep{dmpo2025}               & Diversity             & Token-level diversity objective applied selectively to positive samples. \\
\rowcolor{cTraj!2}
GAPO~\citep{anschel2025gapo}               & Diversity             & Group-aware objective penalises redundant responses. \\
\rowcolor{cTraj!8}
Pro-GRPO~\citep{progrpo2025}               & Diversity             & Over-sample then prune reward-clustered trajectories. \\
\rowcolor{cTraj!2}
DaGRPO~\citep{dagrpo2025}                  & Diversity             & Mask low-distinctiveness pairs, and inject quality anchors. \\
\addlinespace[2pt]
\rowcolor{cTraj!8}
S2R~\citep{s2r2025}                        & Repair                & Self-verification and correction via RL with a two-turn signal. \\
\rowcolor{cTraj!2}
SCoRe~\citep{score2025}                    & Repair                & End-to-end multi-turn self-correction from initial attempt. \\
\rowcolor{cTraj!8}
PAG~\citep{pag2025}                        & Repair                & Policy-as-verifier with revision only on self-detected errors. \\
\midrule
\rowcolor{tabbandTraj}
\multicolumn{3}{l}{\textit{Modifying GRPO update rule}} \\
\rowcolor{cTraj!8}
DAPO~\citep{yu2026dapo}                    & Clip \& ratio         & Asymmetric clip: $\epsilon_{\mathrm{low}} < \epsilon_{\mathrm{high}}$. Also combines dynamic sampling and length shaping (also in Table~\ref{tab:both_variants}). \\
\rowcolor{cTraj!2}
SAPO~\citep{gao2025sapo}                   & Clip \& ratio         & Soft continuous gate replaces hard clip. \\
\rowcolor{cTraj!8}
GSPO~\citep{zheng2025gspo}                 & Clip \& ratio         & Sequence-level IS ratio clipped at sequence level. \\
\bottomrule
\end{tabularx}
\end{table}

\subsection{Reward-Side Variants}
\label{sec:reward_variants}

The central question on the reward side is how to accurately evaluate the quality of each policy decision.
The GRPO advantage $\hat{A}_i$ estimates how much better response $i$ is relative to the group average. Every reward-side intervention ultimately aims to make this estimate more faithful: more informative about which steps contributed to the outcome, less distorted by statistical artifacts of the group-relative formula, and better aligned with the true objective intended by the designer.
Failures in this estimation can occur at every stage of the pipeline: from the quality and granularity of the raw signal $R_i$, through the aggregation and normalization rules that produce $\hat{A}_i$, to the granularity at which the advantage is propagated across token positions.

The reward side encompasses all components that determine $\hat{A}_i$, including the quality of the underlying reward signal $R_i$, the group baseline $\mu_G$, the normalization term $\sigma_G$, and the granularity at which the advantage is assigned.
Reward-side interventions split into two levels: the reward signal $R_i$ itself, and the advantage $\hat{A}_i$ that converts it into a per-token weight.
One class of methods preserves the GRPO advantage formula and instead modifies the reward signal $R_i$ through density, source, shaping, and multi-objective aggregation.
Another class intervenes after $R_i$ is computed, changing how it is transformed into the advantage through statistical bias correction and fine-grained advantage estimation.

\paragraph{Modifying the reward signal $R_i$.}
This design space can be characterized along four dimensions.
Reward density specifies the number of positions in the response that receive a training signal.
Reward source specifies the origin of $R_i$.
Reward shaping specifies which auxiliary terms accompany the primary reward signal.
Multi-objective reward specifies how competing task objectives are aggregated into a single scalar.

\subsubsection*{\textbf{Reward Density}}
\phantomsection\label{sec:reward_density}

\textbf{Problem.} The scalar $R_i$ is typically sparse. It is provided as a single verifier score at the end of the response and does not convey which tokens or reasoning steps within the trajectory contributed to the outcome.
Different task types exhibit qualitatively different sparsity regimes.
Mathematical reasoning often allows binary correctness checking, whereas open-ended generation may provide no directly verifiable signal.
Accordingly, the appropriate design of reward density varies across tasks.

Outcome reward models~\citep{cobbe2021gsm8k} provide the baseline by assigning one scalar reward to each response.
Process reward models~\citep{lightman2024let} increase reward density by labeling intermediate reasoning steps and supplementing or replacing the end-of-response scalar with denser per-step feedback.
GRPO-VPS~\citep{grpovps2026} achieves step-level density without a separately trained reward model.
It segments generation into discrete steps and tracks the conditional probability of the correct answer at each segment boundary, thereby producing verifiable step-level progress signals that are directly incorporated into the GRPO update.
SPO~\citep{guo2025spo} partitions $y_i$ into contiguous segments and assigns a segment-level advantage to each segment, providing intermediate granularity between trajectory-level and token-level credit. Its segment construction also functions as a credit assignment mechanism, which is discussed further in Section~\ref{sec:credit_assign}.
Overall, density-based methods replace the single end-of-response scalar with finer-grained signals, directing the gradient toward the step or segment level where causal responsibility can be localized more precisely.

\subsubsection*{\textbf{Reward Source}}
\phantomsection\label{sec:reward_source}

\textbf{Problem.} The quality of $R_i$ depends on both the source that generates it and the training instances through which it is observed.
Across all three source types, the central reward-side failure is \emph{reward hacking}: the policy optimizes the proxy signal $R_i$ rather than the intended objective, so $R_i$ ceases to track true quality.
Rule-based verifiers are precise within their target domains but brittle outside these domains, where the policy can exploit format, length, or test-case loopholes that satisfy the rule without solving the task.
Trained reward models generalize more effectively but are most susceptible to hacking.
Because they approximate true human preferences from finite data, sustained RL optimization drives the policy toward out-of-distribution outputs that achieve high model scores yet violate the intended objective.
This is a manifestation of Goodhart's Law, which intensifies as the policy diverges from $\piref$~\citep{gao2023scaling}, so the KL penalty toward $\piref$ introduced in Part~I is also the standard structural bound on reward hacking.
LLM-as-judge evaluation~\citep{zheng2023judging} is flexible and scales to open-ended tasks but can be hacked through verbosity or sycophancy and can produce inconsistent results across prompts and models.

Rule-based verifiers are the default choice for mathematical reasoning.
Open-Reasoner-Zero~\citep{openreasoner2025} demonstrates that rule-based verification can support stable training with minimal reward-side complexity.
Image-grounded verifiers extend this paradigm to vision-language settings.
Early vision-language RL systems such as R1-V~\citep{chenr1} indicate that R1-style verifiable-reward training can be extended to multimodal tasks, although verifier design remains task-dependent.
One-shot RLVR~\citep{oneshot_rlvr2025} demonstrates that the data requirement of RLVR can be extremely small when the reward source is precise and verifiable.
With a reliable mathematical verifier, even a single carefully selected training example can yield substantial reasoning gains, indicating that precise verifiable rewards make extremely small RLVR datasets effective.
Overall, the choice of reward source governs both training stability and generalization. 
Rule-based verifiers are reliable within narrow domains. Trained reward models transfer across tasks but may introduce reward hacking. LLM-as-judge methods are flexible but can be inconsistent.

\subsubsection*{\textbf{Reward Shaping}}
\phantomsection\label{sec:reward_shaping}

\textbf{Problem.} The main task reward alone may be insufficient.
It can be insensitive to response length or unaware of when reasoning has already reached a correct conclusion.
Auxiliary reward terms can shape the behavior of the policy along these dimensions.
However, they must be calibrated carefully to avoid displacing the primary objective.

LASER~\citep{laser2025} adds a step-function length reward.
Correct responses within a target-length threshold receive an auxiliary bonus, and a difficulty-aware variant dynamically adapts the threshold while maintaining at least one complete and correct response per difficulty level.
This design achieves a Pareto-optimal balance between correctness and token efficiency.
S-GRPO~\citep{sgrpo2025} assigns rewards that decay with the position at which the model exits reasoning.
The model samples one reasoning path, selects multiple candidate exit points, and receives exponentially decreasing rewards for later correct exits, thereby incentivizing accurate but concise reasoning termination.
Reward shaping is the primary entry point for S-GRPO, but its intervention extends to group construction and advantage computation as well. It is not separately examined in Sections~\ref{sec:stat_bias} or~\ref{sec:credit_assign}.
Overall, shaping methods extend the reward signal to dimensions that the primary verifier ignores. Each auxiliary term introduces a calibration burden, and a weight that is too large can displace the primary objective.

\subsubsection*{\textbf{Multi-Objective Reward}}
\phantomsection\label{sec:multi_obj_reward}

\textbf{Problem.} A single scalar reward $R_i$ cannot simultaneously represent correctness, format adherence, fluency, safety, and user preference unless these objectives are reduced to a weighted sum.
Such a sum requires manually tuned weights, and its components may drive the policy in conflicting directions.
This limitation is structural rather than incidental.
It arises whenever a task requires balancing incommensurable objectives, which is the default setting for general-purpose language models.

Reward aggregation methods construct a composite scalar from heterogeneous reward sources or verifiable components.
QwQ~\citep{qwq2025} combines outcome-based verifiers for mathematics and coding with general reward models and rule-based verifiers in later RL stages.
Kimi k1.5~\citep{kimi2025k15} combines verifiable rewards, learned reward models, and an auxiliary length-based reward that is added to the original reward with an explicit weighting parameter.
GLM-4.5V~\citep{glmv2025} illustrates the engineering complexity of this design at scale. It combines RLVR and RLHF across many multimodal domains and capabilities, using a multi-domain unified reward system that shares common evaluation logic while allowing targeted verifier optimization for each subdomain.
The corresponding training report further observes that weakness in the reward signal for a single capability can disrupt training across domains.
All such approaches modify the quantity that enters $R_i$ while leaving the importance ratio, the clipping rule, and the GRPO advantage formula unchanged.
Overall, multi-objective methods address the structural limitation of a single scalar reward by composing objectives into a calibrated weighted sum, while sharing the challenge of preventing any single component from dominating the gradient.

\begin{table}[!t]
\centering
\caption{Reward-side post-GRPO methods, grouped by whether they modify the reward signal $R_i$ or the advantage $\hat{A}_i$.}
\label{tab:reward_variants}
\scriptsize
\setlength{\tabcolsep}{4pt}
\renewcommand{\arraystretch}{1.0}
\begin{tabularx}{\textwidth}{>{\raggedright\arraybackslash}p{3.0cm}>{\raggedright\arraybackslash}p{2.6cm}X}
\toprule
\rowcolor{tabheadRew}
Method & Dimension & Key mechanism \\
\midrule
\rowcolor{tabbandRew}
\multicolumn{3}{l}{\textit{Modifying the reward signal $R_i$}} \\
\rowcolor{cRew!8}
GRPO-VPS~\citep{grpovps2026}          & Reward density        & Verifiable step-level progress via conditional probability at segment boundaries (also under Credit assignment below). \\
\rowcolor{cRew!2}
SPO~\citep{guo2025spo}                & Reward density        & Segment-level advantage at intermediate granularity between trajectory and token level (also under Credit assignment below). \\
\rowcolor{cRew!8}
Open-Reasoner-Zero~\citep{openreasoner2025} & Reward source   & Rule-based verifier for stable training with minimal complexity. \\
\rowcolor{cRew!2}
R1-V~\citep{chenr1}                   & Reward source         & Image-grounded verifier extends RLVR to vision-language tasks. \\
\rowcolor{cRew!8}
One-shot RLVR~\citep{oneshot_rlvr2025} & Reward source        & Single carefully selected training example suffices when verifier is precise. \\
\rowcolor{cRew!2}
LASER~\citep{laser2025}               & Reward shaping        & Step-function length reward with difficulty-aware dynamic threshold. \\
\rowcolor{cRew!8}
S-GRPO~\citep{sgrpo2025}              & Reward shaping        & Exponentially decaying rewards by reasoning exit position. \\
\rowcolor{cRew!2}
QwQ~\citep{qwq2025}                   & Multi-objective       & Verifiers for math/code combined with reward models across RL stages. \\
\rowcolor{cRew!8}
Kimi k1.5~\citep{kimi2025k15}         & Multi-objective       & Verifiable rewards, learned RM, and length reward with explicit weight. \\
\rowcolor{cRew!2}
GLM-4.5V~\citep{glmv2025}             & Multi-objective       & Multi-domain unified reward system across RLVR and RLHF capabilities. \\
\midrule
\rowcolor{tabbandRew}
\multicolumn{3}{l}{\textit{Modifying the advantage $\hat{A}_i$}} \\
\rowcolor{cRew!8}
Dr.GRPO~\citep{liu2025understanding}  & Stat.\ bias corr.     & Remove per-length and $\sigma_G$ normalization for unbiased estimator. \\
\rowcolor{cRew!2}
Lopti~\citep{yang2025not}            & Stat.\ bias corr.     & Probability-correlated token weights downweight low-prob.\ gradient dominance. \\
\rowcolor{cRew!8}
NGRPO~\citep{nan2025ngrpo}            & Stat.\ bias corr.     & Virtual maximum-reward sample restores nonzero advantages in homogeneous groups. \\
\rowcolor{cRew!2}
AGPO~\citep{li2025agpo}               & Stat.\ bias corr.     & Mask homogeneous groups from loss, and clip negative batch losses to zero. \\
\rowcolor{cRew!8}
CPPO~\citep{cppo2025}                 & Stat.\ bias corr.     & Prune low-absolute-advantage completions before gradient step. \\
\rowcolor{cRew!2}
RiskPO~\citep{ren2025riskpo}          & Stat.\ bias corr.     & Mixed Value-at-Risk baseline prevents entropy collapse from high-prob.\ overemphasis. \\
\rowcolor{cRew!8}
TreeRPO~\citep{yang2025treerpo}       & Credit assignment     & Step-level expected reward estimated from tree-structured rollouts (also under Diversity in Table~\ref{tab:traj_variants}). \\
\rowcolor{cRew!2}
SCRL~\citep{scrl2026}                 & Credit assignment     & Subproblem-level normalized advantages from verifiable reasoning chain decomposition. \\
\rowcolor{cRew!8}
SPO~\citep{guo2025spo}                & Credit assignment     & Segment-level advantage assignment with flexible segmentation scheme. \\
\rowcolor{cRew!2}
HighEnt~\citep{wang2025entropy_token} & Credit assignment     & Selective gradient update on high-entropy forking tokens only. \\
\rowcolor{cRew!8}
PRIME~\citep{cui2025process}          & Credit assignment     & Per-token implicit reward $\beta\log\pi_\phi/\piref$ trained online from outcome labels. \\
\rowcolor{cRew!2}
GRPO-$\lambda$~\citep{grpo_lambda2025} & Credit assignment    & TD($\lambda$)-style eligibility-trace weighting via per-token log-prob. \\
\rowcolor{cRew!8}
VAPO~\citep{vapo2025}                 & Credit assignment     & A value model is reintroduced with decoupled GAE, length-adaptive $\lambda$, asymmetric clipping, and group-sampled contrastive signals to address sparse rewards and length heterogeneity. \\
\rowcolor{cRew!2}
GRPO-VPS~\citep{grpovps2026}          & Credit assignment     & Verifiable step-level progress signals via conditional answer probability at segment boundaries (also under Reward density above). \\
\bottomrule
\end{tabularx}
\end{table}

\paragraph{Modifying the advantage $\hat{A}_i$.}
These methods intervene after $R_i$ has been computed, changing how reward information is transformed into policy-gradient updates.
They can be organized along two dimensions.
Statistical bias correction addresses systematic distortions introduced by the normalization, centering, and aggregation rules used to convert group rewards into advantages and losses.
Credit assignment addresses the uniform broadcast of $\hat{A}_i$ across all token positions, replacing it with a position-sensitive signal intended to better approximate the contribution of each step or token.

\subsubsection*{\textbf{Statistical Bias Correction}}
\phantomsection\label{sec:stat_bias}

\textbf{Problem.} The GRPO advantage formula introduces several sources of systematic distortion, arising from its length averaging, group normalization, group centering, and token-level gradient rules.
The per-response length averaging in standard GRPO scales each response-level loss by $1/|y_i|$, so long incorrect responses receive a smaller per-token penalty than short incorrect responses, creating a length-dependent aggregation bias.
Normalization by $\sigma_G$ amplifies groups with small within-group reward variance, making the update scale depend on sampled group statistics rather than only on task utility.
When all $G$ responses in a group receive identical rewards, both the centered numerator $R_i-\mu_G$ and the denominator $\sigma_G$ vanish, so the normalized advantage is formally $0/0$. Implementations resolve this by adding a small constant $\varepsilon$ to $\sigma_G$ or by skipping the group, which sets the advantage to zero and blocks the gradient entirely, both on easy prompts where all responses are correct and on hard prompts where all are incorrect.

Group-mean centering also discards absolute reward scale: groups with different absolute success rates can yield similar normalized advantages if their relative reward patterns are similar.
At the token level, low-probability tokens can produce disproportionately large gradient magnitudes and dominate the update even when they are not semantically responsible for the final outcome.
These effects arise from particular normalization and aggregation choices in GRPO-style training rather than from verifiable rewards themselves, and they can accumulate over the course of RL training.

Dr.~GRPO~\citep{liu2025understanding} removes both response-length normalization and $\sigma_G$ normalization to reduce the length and difficulty biases of vanilla GRPO, trading these bias reductions for higher gradient variance.
Lopti~\citep{yang2025not} assigns token-level loss weights that are positively correlated with the predicted probability of each token, downweighting low-probability tokens whose large gradient magnitudes would otherwise dominate GRPO updates.
NGRPO~\citep{nan2025ngrpo} hypothesizes a virtual maximum-reward sample during advantage computation, altering the group mean and variance so that homogeneously incorrect responses receive nonzero advantages rather than being silently zeroed out.

AGPO~\citep{li2025agpo} masks zero-advantage groups from the batch loss computation, preventing all-correct or all-wrong groups from entering the mean loss as uninformative terms, and clips negative batch losses to zero to mitigate overconfidence and entropy collapse.
CPPO~\citep{cppo2025} prunes completions with low absolute advantages before computing the gradient, reducing the number of responses that enter the loss while preserving the signal from high-advantage completions.
RiskPO~\citep{ren2025riskpo} replaces mean-based optimization with a Mixed Value-at-Risk objective that integrates weighted attention over multiple regions of the reward distribution, mitigating entropy collapse and overconfident convergence caused by overemphasis on high-probability sequences.
Overall, these methods mitigate statistical artifacts induced by GRPO's normalization and aggregation choices, each targeting a distinct distortion: length bias, variance amplification, degenerate advantages, or token-level gradient dominance.

\subsubsection*{\textbf{Credit Assignment and Fine-Grained Advantage}}
\phantomsection\label{sec:credit_assign}

\textbf{Problem.} The credit assignment problem in RL asks which actions are causally responsible for the observed return.
In GRPO, this problem takes a specific form: the scalar $\hat{A}_i$ is broadcast uniformly to every token in a response, treating all positions as equally responsible for the outcome.
This uniform assignment wastes gradient on uninformative tokens and dilutes the signal at decisive ones.
The natural resolution is to replace the uniform broadcast with a position-sensitive signal, either by structurally decomposing the reward across steps, by selectively updating high-signal positions, or by estimating a fine-grained advantage directly from model internals.

At the step level via structural decomposition, TreeRPO~\citep{yang2025treerpo} samples a tree of continuations from intermediate reasoning states and estimates step-level expected rewards from the resulting tree-structured rollouts.
SCRL~\citep{scrl2026} decomposes a reference reasoning chain into verifiable subproblems and assigns subproblem-level normalized advantages to the corresponding answer segments, providing a finer-grained credit signal than a single end-of-response scalar.
SPO~\citep{guo2025spo} partitions $y_i$ into contiguous segments under a flexible segmentation scheme and assigns a segment-level advantage to each segment, providing an intermediate granularity between trajectory-level and token-level credit assignment.

Via selective updating, HighEnt~\citep{wang2025entropy_token} shows that only a small fraction of tokens exhibit high entropy and that these high-entropy tokens act as forking tokens that steer the model toward qualitatively different reasoning paths.
Applying policy gradient updates selectively to this minority maintains or improves the performance of full-gradient training, whereas updating only the lowest-entropy tokens degrades performance.

Via model-driven estimation, PRIME~\citep{cui2025process} derives per-token implicit process rewards as $\beta \log \frac{\pi_\phi(y_t \mid y_{<t})}{\piref(y_t \mid y_{<t})}$, where $\pi_\phi$ is a language model trained online from outcome labels and $\piref$ is a fixed reference, producing a dense token-level signal without requiring explicit step-level annotations.
GRPO-$\lambda$~\citep{grpo_lambda2025} approximates a TD($\lambda$)-style advantage by reformulating eligibility-trace weighting through per-token log-probabilities, enabling fine-grained credit assignment across token positions without a learned value model.

Although broader than pure credit assignment, VAPO~\citep{vapo2025} revisits advantage estimation through a value-based framework. Value model bias is corrected by value pre-training and a decoupled GAE that uses separate $\lambda$ values for policy and value updates. Heterogeneous sequence lengths are handled by dynamically adjusting $\lambda$ according to sequence length. Sparse rewards are addressed jointly through asymmetric clipping, positive example imitation loss, and group-sampled contrastive signals. In reintroducing a value model, VAPO partially converges with the PPO approach, applying the value model selectively to resolve limitations that the group-relative baseline alone cannot address.
Also relevant here, GRPO-VPS~\citep{grpovps2026} produces verifiable step-level progress signals by tracking conditional answer probability at segment boundaries, which serves simultaneously as a density mechanism and a credit signal.
Overall, credit assignment methods break the uniform-broadcast assumption of GRPO and concentrate the gradient on tokens or steps that are more likely to carry useful learning signal, whether through structural decomposition, selective updating, or model-driven estimation.

\subsection{Both-Side Variants: Compound Instabilities}
\label{sec:both_variants}

Some failure modes are better understood as interactions between the trajectory side and the reward side. Modifying one side may leave residual instability on the other.
Existing both-side methods typically assemble independently motivated fixes rather than deriving them from a shared principle. The open gaps below highlight couplings that remain underexplored or have not yet been systematically formulated.

\begin{table}[!tp]
\centering
\caption{Both-side post-GRPO methods for compound instabilities coupling the two sides.}
\label{tab:both_variants}
\scriptsize
\setlength{\tabcolsep}{4pt}
\renewcommand{\arraystretch}{1.05}
\begin{tabularx}{\textwidth}{>{\raggedright\arraybackslash}p{3.0cm}>{\raggedright\arraybackslash}p{2.6cm}X}
\toprule
\rowcolor{tabheadBoth}
Method & Dimension & Key mechanism \\
\midrule
\rowcolor{cBoth!8}
DAPO~\citep{yu2026dapo}               & Clip--variance        & Asymmetric clipping on the trajectory side is combined with dynamic sampling, token-level policy-gradient loss, and length shaping on the reward side (also under Clip \& ratio in Table~\ref{tab:traj_variants}). \\
\rowcolor{cBoth!2}
DCPO~\citep{yang2025dcpo}             & Clip--variance        & Token-specific dynamic clip thresholds on the trajectory side are paired with smooth advantage standardization over cumulative training statistics on the reward side. \\
\addlinespace[2pt]
\rowcolor{cBoth!8}
HT-GRPO~\citep{htgrpo2025}            & Granularity           & A prompt-conditioned ratio estimator on the trajectory side is coupled with phase-differentiated advantage weighting on the reward side. \\
\rowcolor{cBoth!2}
GTPO~\citep{gtpo2025}                 & Granularity           & Policy entropy on the trajectory side defines the token-level granularity at which the signal is applied, paired with entropy-weighted token-level reward redistribution on the reward side. \\
\bottomrule
\end{tabularx}
\end{table}

\subsubsection*{\textbf{Clip--Variance Coupling}}

\textbf{Problem.}
Symmetric clipping and group-relative normalization can interact adversely during long-CoT RL.
If clipping suppresses exploration and the sampled group becomes increasingly homogeneous, the within-group reward variance $\sigma_G$ can shrink or vanish, reducing the usefulness of group-relative advantages.
Conversely, variance-based normalization can make update scales highly dependent on sampled group statistics. As a result, trajectory-side clipping and reward-side normalization may reinforce instability rather than correct it independently.

DAPO~\citep{yu2026dapo} combines Clip-Higher, an asymmetric clipping rule with $\epsilon_{\mathrm{low}} < \epsilon_{\mathrm{high}}$, with dynamic sampling, token-level policy-gradient loss, and overlong reward shaping.
Each component addresses a distinct failure mode. DAPO shows that these independently motivated components can be complementary in a practical recipe for stable long-CoT training.
DCPO~\citep{yang2025dcpo} uses token-specific dynamic clip thresholds on the trajectory side and smooth advantage standardization over cumulative training statistics on the reward side, reducing the zero-gradient problem that arises when all responses in a group receive identical rewards.

\subsubsection*{\textbf{Granularity Coupling}}

\textbf{Problem.}
Ratio granularity and advantage granularity are not independent.
A token-level ratio applied to a step-level advantage can mismatch the unit of trust-region control with the unit of credit.
A sequence-level ratio applied to a token-level advantage can create the reverse mismatch.
A principled design should therefore specify ratio aggregation and advantage assignment at compatible granularities.

GTPO~\citep{gtpo2025} is closer to the reward-side boundary of this category: it uses policy entropy to define token-level reward or advantage weighting, concentrating the learning signal on uncertain positions rather than broadcasting the same scalar uniformly.
This differs from HighEnt in Section~\ref{sec:credit_assign}, which selectively applies policy-gradient updates to high-entropy tokens, whereas GTPO redistributes token-level reward or advantage weights according to entropy.
Although most GRPO variants discussed here target text-only long-CoT RLVR, similar coupling patterns also appear in adjacent generation settings.
Sketch-Then-Paint / HT-GRPO~\citep{htgrpo2025} anchors the importance ratio to a prompt-conditioned estimator computed from a fully masked state on the trajectory side and assigns advantage weights by the structural role of each token in the staged generation process on the reward side.
The two choices are coupled through the stage structure. The stages organize the masked-state ratio computation and define the structural roles used for advantage weighting.
Both methods rely on method-specific anchors, such as a phase boundary in HT-GRPO or a policy-entropy estimate in GTPO. A general joint granularity design without such an anchor remains open.

\subsubsection*{\textbf{Open Gaps in Both-Side Design}}

The methods above address specific instabilities, but two deeper gaps persist across Part~II.

\textbf{Absence of a joint theoretical foundation.}
The objective $J(\theta)=\mathbb{E}_{\tau\sim p_\theta}[R(\tau)]$ couples the trajectory distribution $p_\theta(\tau)$ with the reward functional $R(\tau)$. However, existing GRPO/RLVR variants rarely derive from this structure the conditions under which trajectory-side and reward-side choices should co-vary.
Without such a theory, it is difficult to determine from first principles whether a single-side fix is sufficient or a joint intervention is necessary.
The both-side methods reviewed above largely assemble components that were motivated independently on each side rather than derived from a shared principle.

\textbf{Uncharted interactions between single-side methods.}
Trajectory-side and reward-side methods operate simultaneously within the same update step.
A trajectory-side method that changes the sample distribution also changes the groups on which reward normalization, filtering, or shaping operates. Conversely, reward-side filtering or weighting changes which trajectories contribute to the effective policy update.
These interactions have not been systematically studied, so the combined behavior of methods drawn from different parts of Part~II remains unknown.
Combining independently validated methods does not guarantee a well-behaved joint system.

\part*{Part~III\quad Structural Extensions: Agentic RL and On-Policy Distillation}
\addcontentsline{toc}{part}{Part III: Structural Extensions: Agentic RL and On-Policy Distillation}

\section{Agentic RL as a Trajectory-Side Structural Extension}
\label{sec:agentic}

Part~II showed that GRPO originates as a reward-side substitution and that subsequent work refines the same single-turn estimator on both sides, while remaining within the regime of single-turn token generation.
Part~III departs structurally from that endpoint in two complementary directions.
This section addresses the first direction: Agentic RL, which expands \emph{what the trajectory contains} while preserving the GRPO update rule and reward aggregation logic.
The action space expands from token generation against a static prompt to structured multi-step decisions, including web search, code execution, and GUI interaction, interleaved with environment observations over multiple turns.
The introduction of an environment constitutes the single structural change that induces all new failure modes on both sides.
Section~\ref{sec:opd_line} addresses the second direction.

A note on scope: Agentic RL is an active research area with a large and rapidly growing body of work.
Rather than attempting to catalog this literature comprehensively, this section uses the trajectory-side and reward-side framework developed in Part~II as an analytical lens, identifying which Part~II failure modes carry over structurally and which new failure modes are introduced by the environment.
The goal is to show how the same analytical axes extend to the agentic setting, not to provide a complete survey of agentic methods.
This focus complements dedicated agentic RL surveys. The landscape survey of \citet{zhang2025landscape} organizes the field by agentic capability, such as planning, tool use, and memory, and catalogs what agents can do. Domain surveys of deep research systems take the same capability-centric view within one application~\citep{li2025rldrsurvey,zhang2025agenticdr}. The two-sided reading is orthogonal. It asks instead which factor of $\nabla_\theta J(\theta)$ a method modifies and which failure that change answers. The two readings compose. A capability such as memory or tool use is ultimately realized as a trajectory-side or reward-side intervention on the same estimator.

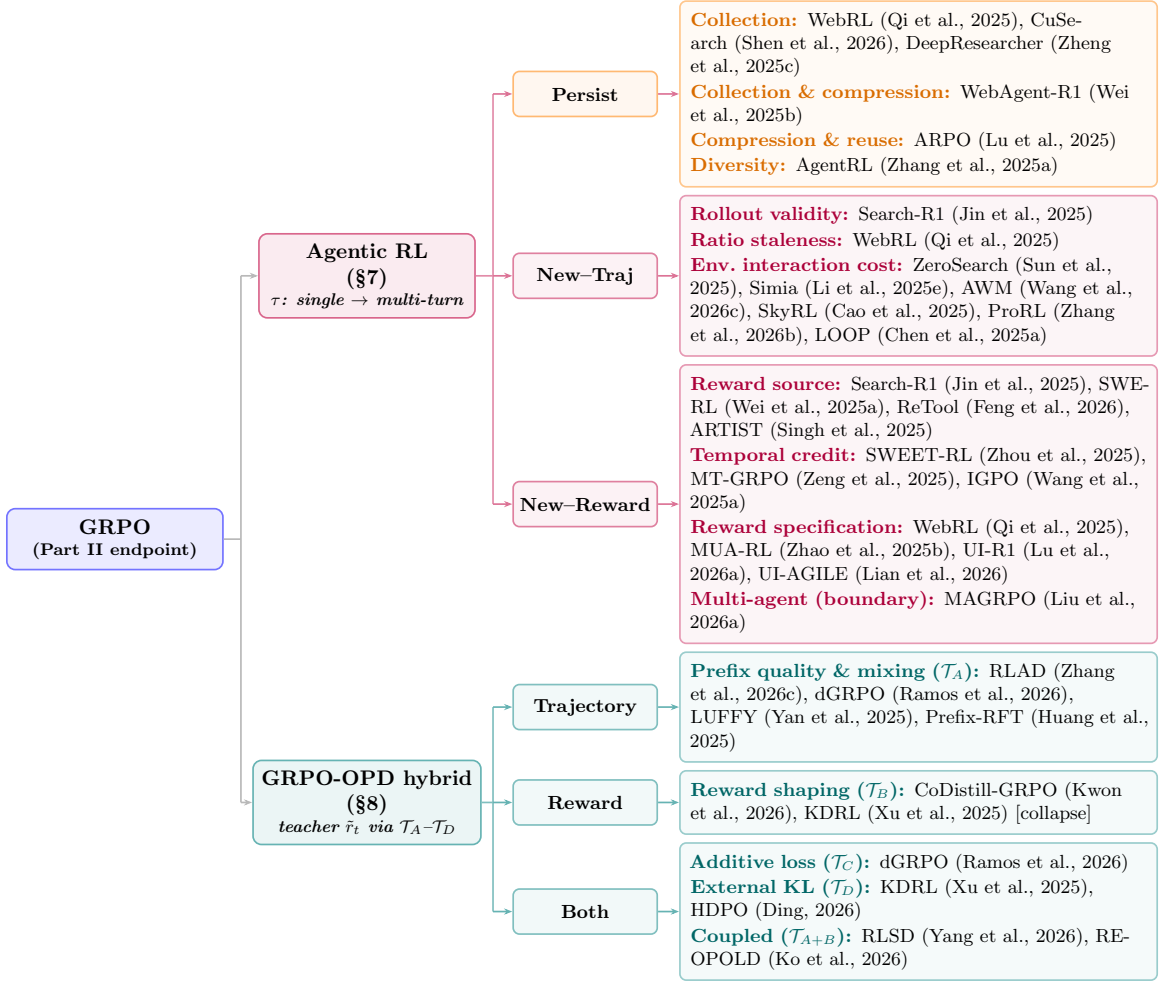
\begin{figure}[t]
\centering
\resizebox{\textwidth}{!}{%
\begin{tikzpicture}[
  x=1cm, y=0.65cm,
  ROOT/.style={
    draw=blue!55, fill=blue!8, rounded corners=4pt, thick,
    font=\small\bfseries, align=center,
    minimum width=3.6cm, minimum height=1.0cm, inner sep=5pt
  },
  BRANCH/.style={
    rounded corners=4pt, thick,
    font=\small\bfseries, align=center,
    minimum width=3.6cm, minimum height=1.4cm, inner sep=4pt
  },
  SUBBRANCH/.style={
    rounded corners=3pt, thick,
    font=\footnotesize\bfseries, align=center,
    minimum width=2.4cm, minimum height=0.75cm, inner sep=3pt
  },
  LEAF/.style={
    rounded corners=3pt, thick,
    font=\footnotesize, align=left,
    text width=7.6cm, inner sep=5pt
  },
  gline/.style={thick, gray!55},
  garr/.style={-{Stealth[length=4pt,width=3pt]}, thick, gray!55},
  parr/.style={-{Stealth[length=4pt,width=3pt]}, thick, purple!55},
  pline/.style={thick, purple!45},
  tarr/.style={-{Stealth[length=4pt,width=3pt]}, thick, teal!60},
  tline/.style={thick, teal!50},
]

\def\leafgap{0.12cm}

\def\subleafgap{0.35cm}


\node[LEAF, draw=cTraj!45, fill=cTraj!4, anchor=west] (lag_p) at (10.8, 5.366) {%
  \textcolor{cTraj!85!black}{\textbf{Collection:}} WebRL~\citep{qi2024webrl}, CuSearch~\citep{shen2026cusearch},
    DeepResearcher~\citep{zheng2025deepresearcher}\\[1pt]
  \textcolor{cTraj!85!black}{\textbf{Collection \& compression:}} WebAgent-R1~\citep{wei2025webagentr1}\\[1pt]
  \textcolor{cTraj!85!black}{\textbf{Compression \& reuse:}} ARPO~\citep{lu2025arpo}\\[1pt]
  \textcolor{cTraj!85!black}{\textbf{Diversity:}} AgentRL~\citep{zhang2025agentrl}
};

\node[LEAF, draw=purple!40, fill=purple!4, anchor=north west] (lag_t)
  at ([yshift=-\leafgap]lag_p.south west) {%
  \textcolor{cAgent!90!black}{\textbf{Rollout validity:}} Search-R1~\citep{jin2025search}\\[1pt]
  \textcolor{cAgent!90!black}{\textbf{Ratio staleness:}} WebRL~\citep{qi2024webrl}\\[1pt]
  \textcolor{cAgent!90!black}{\textbf{Env.\ interaction cost:}} ZeroSearch~\citep{zerosearch2025}, Simia~\citep{li2025simia},
    AWM~\citep{wang2026awm}, SkyRL~\citep{cao2025skyrl}, ProRL~\citep{zhang2026prorl},
    LOOP~\citep{chen2025loop}
};

\node[LEAF, draw=purple!40, fill=purple!4, anchor=north west] (lag_r)
  at ([yshift=-\leafgap]lag_t.south west) {%
  \textcolor{cAgent!90!black}{\textbf{Reward source:}} Search-R1~\citep{jin2025search}, SWE-RL~\citep{wei2026swe},
    ReTool~\citep{feng2025retool}, ARTIST~\citep{singh2025artist}\\[1pt]
  \textcolor{cAgent!90!black}{\textbf{Temporal credit:}} SWEET-RL~\citep{zhou2025sweetrl}, MT-GRPO~\citep{wei2025mtgrpo},
    IGPO~\citep{wang2025igpo}\\[1pt]
  \textcolor{cAgent!90!black}{\textbf{Reward specification:}} WebRL~\citep{qi2024webrl}, MUA-RL~\citep{zhao2025muarl},
    UI-R1~\citep{lu2025uir1}, UI-AGILE~\citep{lian2025uiagile}\\[1pt]
  \textcolor{cAgent!90!black}{\textbf{Multi-agent (boundary):}} MAGRPO~\citep{liu2025magrpo}
};

\node[LEAF, draw=teal!40, fill=teal!4, anchor=north west] (lopd_t)
  at ([yshift=-\leafgap]lag_r.south west) {%
  \textcolor{cOPD!80!black}{\textbf{Prefix quality \& mixing ($\mathcal{T}_A$):}} RLAD~\citep{rlad2025}, dGRPO~\citep{ramos2026combining}, LUFFY~\citep{yan2025luffy}, Prefix-RFT~\citep{huang2025prefixrft}
};

\node[LEAF, draw=teal!40, fill=teal!4, anchor=north west] (lopd_r)
  at ([yshift=-\leafgap]lopd_t.south west) {%
  \textcolor{cOPD!80!black}{\textbf{Reward shaping ($\mathcal{T}_B$):}} CoDistill-GRPO~\citep{codistill_grpo2025}, KDRL~\citep{xu2025kdrl} [collapse]
};

\node[LEAF, draw=teal!40, fill=teal!4, anchor=north west] (lopd_b)
  at ([yshift=-\leafgap]lopd_r.south west) {%
  \textcolor{cOPD!80!black}{\textbf{Additive loss ($\mathcal{T}_C$):}} dGRPO~\citep{ramos2026combining}\\[1pt]
  \textcolor{cOPD!80!black}{\textbf{External KL ($\mathcal{T}_D$):}} KDRL~\citep{xu2025kdrl}, HDPO~\citep{hdpo2025}\\[1pt]
  \textcolor{cOPD!80!black}{\textbf{Coupled ($\mathcal{T}_{A{+}B}$):}} RLSD~\citep{rlsd2025}, REOPOLD~\citep{reopold2025}
};


\node[SUBBRANCH, draw=cTraj!55, fill=cTraj!6, anchor=east]
  (ag_p) at ([xshift=-\subleafgap]lag_p.west) {Persist};

\node[SUBBRANCH, draw=purple!50, fill=purple!5, anchor=east]
  (ag_t) at ([xshift=-\subleafgap]lag_t.west) {New--Traj};

\node[SUBBRANCH, draw=purple!50, fill=purple!5, anchor=east]
  (ag_r) at ([xshift=-\subleafgap]lag_r.west) {New--Reward};

\node[SUBBRANCH, draw=teal!50, fill=teal!5, anchor=east]
  (opd_t) at ([xshift=-\subleafgap]lopd_t.west) {Trajectory};

\node[SUBBRANCH, draw=teal!50, fill=teal!5, anchor=east]
  (opd_r) at ([xshift=-\subleafgap]lopd_r.west) {Reward};

\node[SUBBRANCH, draw=teal!50, fill=teal!5, anchor=east]
  (opd_b) at ([xshift=-\subleafgap]lopd_b.west) {Both};


\node[BRANCH, draw=purple!60, fill=purple!8] (ag)
  at (5.6, 0 |- ag_t)
  {Agentic RL\\(\S\ref{sec:agentic})\\[-2pt]\scriptsize\itshape $\tau$: single $\to$ multi-turn};

\node[BRANCH, draw=teal!60, fill=teal!8] (opd)
  at (5.6, 0 |- opd_r)
  {GRPO-OPD hybrid\\(\S\ref{sec:opd_line})\\[-2pt]\scriptsize\itshape teacher $\tilde{r}_t$ via $\mathcal{T}_A$--$\mathcal{T}_D$};


\node[ROOT] (grpo) at ($(1.4,0 |- ag)!0.5!(1.4,0 |- opd)$)
  {GRPO\\[-2pt]\scriptsize(Part~II endpoint)};


\coordinate (fork) at (3.5, 0 |- grpo);

\draw[gline] (grpo.east) -- (fork);
\draw[gline] (fork |- ag) -- (fork |- opd);
\draw[garr]  (fork |- ag)   -- (ag.west);
\draw[garr]  (fork |- opd)  -- (opd.west);


\coordinate (agfork) at (7.7, 0 |- ag);

\draw[pline] (ag.east) -- (agfork);
\draw[pline] (agfork |- ag_p) -- (agfork |- ag_r);
\draw[parr]  (agfork |- ag_p) -- (ag_p.west);
\draw[parr]  (agfork |- ag_t) -- (ag_t.west);
\draw[parr]  (agfork |- ag_r) -- (ag_r.west);


\coordinate (opdfork) at (7.7, 0 |- opd);

\draw[tline] (opd.east) -- (opdfork);
\draw[tline] (opdfork |- opd_t) -- (opdfork |- opd_b);
\draw[tarr]  (opdfork |- opd_t) -- (opd_t.west);
\draw[tarr]  (opdfork |- opd_r) -- (opd_r.west);
\draw[tarr]  (opdfork |- opd_b) -- (opd_b.west);


\draw[parr] (ag_p.east)  -- (lag_p.west);
\draw[parr] (ag_t.east)  -- (lag_t.west);
\draw[parr] (ag_r.east)  -- (lag_r.west);

\draw[tarr] (opd_t.east) -- (lopd_t.west);
\draw[tarr] (opd_r.east) -- (lopd_r.west);
\draw[tarr] (opd_b.east) -- (lopd_b.west);

\end{tikzpicture}%
}

\caption{%
  Part~III structural extensions from the Part~II GRPO endpoint.
  Each extension alters one axis of $J(\theta)=\mathbb{E}_{\tau\sim p_\theta}[R(\tau)]$.
  \textbf{Agentic RL} expands $\tau$ to multi-turn environment interaction.
  \textbf{The GRPO-OPD hybrid} retains the GRPO update rule and introduces the teacher signal $\tilde{r}_t$
  along the same trajectory-side, reward-side, and both-side axes used in Part~II.
  Operator labels $\mathcal{T}_A$--$\mathcal{T}_D$ indicate where each method intervenes.
  The orange node in the Agentic branch lists Part~II problem dimensions that persist there.
  Colored nodes list qualitatively new dimensions introduced by the structural change.
}
\label{fig:part3_overview}
\end{figure}

\paragraph{Formal objective.}
Agentic RL replaces the single-turn response $y_i$ with a multi-turn interaction sequence:
\begin{equation}
  \label{eq:agentic_traj}
  \tau_i = (x,\; o_{i,1},\, a_{i,1},\; o_{i,2},\, a_{i,2},\; \ldots,\; o_{i,H},\, a_{i,H}),
\end{equation}
where $o_{i,t}$ denotes the environment observation at turn $t$ and $a_{i,t}$ denotes the action of the model.
Substituting this trajectory into the GRPO objective yields:
\begin{equation}
  \label{eq:agentic_grpo}
  \begin{split}
  L^{\mathrm{Agentic}}(\theta)
  ={} & \E\!\left[
      \frac{1}{G}\sum_{i=1}^{G}
      \frac{1}{|\tau_i^{\mathrm{mask}}|}
      \sum_{t=1}^{H}\sum_{k=1}^{|\tau_{i,t}^{\mathrm{mask}}|}
      \min\!\bigl(r_{i,t,k}\,\hat{A}_i,\;
                  \clip(r_{i,t,k},1{-}\epsilon,1{+}\epsilon)\,\hat{A}_i\bigr)
    \right] \\
  &\quad\quad\quad\quad\quad\quad\quad\quad\quad\quad\quad\quad\quad\quad\quad\quad\quad\quad\quad\ - \beta\,\KL(\pitheta\|\piref),
  \end{split}
\end{equation}
where $\tau_{i,t}^{\mathrm{mask}}$ denotes the subset of policy-generated tokens at turn $t$ that enter the policy-gradient loss, $|\tau_i^{\mathrm{mask}}| = \sum_{t=1}^{H}|\tau_{i,t}^{\mathrm{mask}}|$ is the total number of such tokens in the trajectory, $k$ indexes these tokens within the turn, and $r_{i,t,k}$ is the per-token importance ratio of the $k$-th such token, defined exactly as in single-turn GRPO.
In tool-augmented trajectories, environment-generated observation tokens must be excluded from the likelihood-ratio and KL terms, because the policy did not generate them and assigning gradient credit to them violates the policy-gradient derivation.
Search-R1~\citep{jin2025search} makes this requirement explicit through retrieved-token masking, which excludes retrieved tokens from the policy-gradient loss and restricts updates to policy-generated tokens.
There are two structural modifications relative to single-turn GRPO.
First, the single sum over response tokens becomes a double sum over turns and within-turn action tokens.
Second, $R_i$ is computed from trajectory-level environment feedback, often as a terminal outcome reward, rather than from a single-turn static verifier.

\paragraph{Trajectory side.}
\textbf{How do trajectory-side failures from Part~II persist in Agentic RL?}
Part~II identifies trajectory-side failures at two levels. The first concerns the trajectories themselves, including rollout collection, trajectory compression and reuse, group diversity, and the repair of failed rollouts. The second concerns the update rule, where hard symmetric clipping combined with group-relative normalization can drive entropy collapse, and token-level importance ratios create a granularity mismatch when applied to sequence-level advantages.
These failures persist in the agentic setting. The methods below are illustrative rather than exhaustive. Each is included only to show that a Part~II failure mode reappears in the agentic setting and is addressed by a recognizably trajectory-side intervention.

For trajectory collection, WebRL~\citep{qi2024webrl} uses a self-evolving online curriculum together with outcome-supervised reward modeling and adaptive RL strategies, so that training focuses on tasks that remain informative under the current policy.
CuSearch~\citep{shen2026cusearch} reallocates the rollout budget toward deeper-search trajectories through a search-depth greedy allocation operator, concentrating the gradient signal on trajectories with more retrieval decision points.
WebAgent-R1~\citep{wei2025webagentr1} extends GRPO to the multi-turn web setting and adds context compression and asynchronous rollout. This is a trajectory-side response to the collection and compression problems carried over from Part~II. DeepResearcher~\citep{zheng2025deepresearcher} scales rollout collection in live web-search environments, where the environment rather than a static prompt set decides which trajectories appear.
For compression and reuse, ARPO~\citep{lu2025arpo} maintains a per-task experience replay buffer that substitutes one failed trajectory with a previously successful one when the entire group fails, preventing zero-reward groups from producing vanishing gradients under sparse GUI task rewards. 
It also applies a task selection strategy that retains only tasks on which a baseline agent succeeds at least once, thereby filtering out unlearnable interactions.
For diversity, AgentRL~\citep{zhang2025agentrl} applies Cross-Policy Sampling, drawing actions from multiple policies to encourage exploration in multi-turn settings.

\textbf{What new trajectory-side failures are introduced by interactive environments?}
Beyond these inherited failures, the introduction of a real interactive environment creates three qualitatively new trajectory-side problems with no counterpart in single-turn generation.

\emph{Rollout validity}: a multi-turn trajectory interleaves policy-generated action tokens with observation tokens returned by the environment.
The policy-gradient loss must therefore be restricted to policy-generated tokens.
Including environment-generated observation tokens would incorrectly assign credit to tokens that the policy did not generate, and Search-R1~\citep{jin2025search} establishes retrieved-token masking as the standard remedy for this case.

\emph{Importance ratio staleness}: as the horizon $H$ grows, the environment state distribution shifts after each gradient step, which invalidates importance ratios computed under $\piold$.
WebRL~\citep{qi2024webrl} addresses the related problem of policy distribution drift through a self-evolving online curriculum, outcome-supervised reward modeling, and adaptive RL strategies that update the training distribution as the policy changes.

\emph{Environment cost and availability}: each tool call blocks the rollout until the environment responds, whether the call involves a live search query, a code-execution sandbox, or a GUI action.
This single source of friction surfaces as three related bottlenecks: the per-call cost of each interaction, the availability of an environment to interact with at all, and the throughput of collecting rollouts in parallel.
Wall-clock time per gradient step therefore increases with both the horizon $H$ and the group size $G$, while quota constraints make parallel rollout difficult and directly limit the gradient update budget.
For per-call cost, ZeroSearch~\citep{zerosearch2025} converts the LLM into a retrieval module that simulates search, generating both useful and noisy documents to replace live search calls and eliminate API cost.
For availability, when no real environment exists rollout collection is impossible. Simia~\citep{li2025simia} uses LLM-simulated feedback to support training without real environments, and AWM~\citep{wang2026awm} constructs fully synthetic code-driven and database-backed environments at scale.
For throughput, sequential tool calls block gradient updates and limit parallelism. SkyRL~\citep{cao2025skyrl} provides an efficient multi-turn tool-use RL pipeline with asynchronous rollout, and ProRL~\citep{zhang2026prorl} fully decouples rollout generation as a rollout-as-a-service infrastructure, separating the rollout process from the training loop.

A related limit appears when the horizon grows so long that storing the full trajectory becomes the binding memory constraint. LOOP~\citep{chen2025loop} answers with a memory-efficient PPO variant that keeps one copy of the policy and uses no separate value network. This shares the critic-free motivation of the GRPO line.

\begin{table}[!htp]
\centering
\caption{Agentic RL methods grouped by problem dimension: dimensions where Part~II problems persist, and qualitatively new dimensions introduced by environment interaction. The final row is a multi-agent boundary case.}
\label{tab:agentic_variants}
\scriptsize
\setlength{\tabcolsep}{4pt}
\renewcommand{\arraystretch}{0.98}
\begin{tabularx}{\textwidth}{>{\raggedright\arraybackslash}p{3.0cm}>{\raggedright\arraybackslash}p{2.6cm}X}
\toprule
\rowcolor{tabheadAG}
Method & Dimension & Key mechanism \\
\midrule
\rowcolor{tabbandAG}
\multicolumn{3}{l}{\textit{Part~II problems persist}} \\
\rowcolor{cAgent!7}
Search-R1~\citep{jin2025search}            & Collection / validity   & Retrieved-token masking restricts the policy-gradient loss to policy-generated action tokens. \\
\rowcolor{cAgent!2}
WebRL~\citep{qi2024webrl}                  & Collection              & A self-evolving curriculum filters uninformative task-state pairs to keep group rewards non-degenerate. \\
\rowcolor{cAgent!7}
CuSearch~\citep{shen2026cusearch}          & Collection              & A search-depth greedy allocation operator concentrates the rollout budget on deeper-search trajectories. \\
\rowcolor{cAgent!2}
WebAgent-R1~\citep{wei2025webagentr1}      & Collection \& compr.    & Multi-turn GRPO with context compression and asynchronous rollout. \\
\rowcolor{cAgent!7}
DeepResearcher~\citep{zheng2025deepresearcher} & Collection          & Rollout collection scaled in live web-search environments rather than a static prompt set. \\
\rowcolor{cAgent!2}
ARPO~\citep{lu2025arpo}                    & Compression \& reuse    & A per-task experience replay buffer substitutes one failed trajectory with a previously successful one when the entire group fails, preventing zero-reward groups. \\
\rowcolor{cAgent!7}
AgentRL~\citep{zhang2025agentrl}           & Diversity               & Cross-Policy Sampling draws actions from multiple policies to increase group exploration diversity. \\
\rowcolor{cAgent!2}
SWE-RL~\citep{wei2026swe}                  & Reward source           & A patch-similarity score against the ground-truth patch provides a continuous graded terminal reward. \\
\rowcolor{cAgent!7}
ReTool~\citep{feng2025retool}              & Reward source           & An outcome reward is derived from whether tool calls ultimately produce correct results, rewarding successful tool-use strategies without step-level annotation. \\
\rowcolor{cAgent!2}
ARTIST~\citep{singh2025artist}             & Reward source           & An outcome reward shapes interleaved reasoning and tool calls without step-level supervision. \\
\midrule
\rowcolor{tabbandAG}
\multicolumn{3}{l}{\textit{New problems from environment interaction}} \\
\rowcolor{cAgent!7}
Search-R1~\citep{jin2025search}            & Rollout validity        & Retrieved-token masking excludes environment-generated observation tokens from the policy-gradient loss. \\
\rowcolor{cAgent!2}
WebRL~\citep{qi2024webrl}                  & Ratio staleness         & A self-evolving online curriculum with adaptive RL updates the training distribution as the policy drifts, keeping importance ratios from going stale. \\
\rowcolor{cAgent!7}
ZeroSearch~\citep{zerosearch2025}          & Env.\ interaction cost  & An LM-based retrieval simulator replaces live search calls and decouples rollout throughput from API availability. \\
\rowcolor{cAgent!2}
Simia~\citep{li2025simia}                  & Env.\ interaction cost  & An LLM-simulated environment provides feedback for training without access to a real environment. \\
\rowcolor{cAgent!7}
AWM~\citep{wang2026awm}                    & Env.\ interaction cost  & Procedurally generated synthetic environments remove any dependency on external services. \\
\rowcolor{cAgent!2}
SkyRL~\citep{cao2025skyrl}                 & Env.\ interaction cost  & Asynchronous rollout dispatch reduces wall-clock time per gradient step in multi-turn training. \\
\rowcolor{cAgent!7}
ProRL~\citep{zhang2026prorl}               & Env.\ interaction cost  & A rollout-as-a-service infrastructure fully decouples rollout generation from the training loop. \\
\rowcolor{cAgent!2}
LOOP~\citep{chen2025loop}                  & Env.\ cost (memory)     & A memory-efficient PPO variant keeps one policy copy and no value network for long horizons. \\
\rowcolor{cAgent!7}
SWEET-RL~\citep{zhou2025sweetrl}           & Temporal credit         & An auxiliary model scores partial trajectory prefixes to produce per-turn advantage estimates. \\
\rowcolor{cAgent!2}
MT-GRPO~\citep{wei2025mtgrpo}              & Temporal credit         & Turn-level rewards blend each turn's feedback with a share of the final advantage. \\
\rowcolor{cAgent!7}
IGPO~\citep{wang2025igpo}                  & Temporal credit         & Per-turn information gain from the model's belief update gives dense supervision without a reward model. \\
\rowcolor{cAgent!2}
WebRL~\citep{qi2024webrl}                  & Reward specification    & An outcome-supervised reward model scores final web-task success and feeds a self-evolving curriculum for online training. \\
\rowcolor{cAgent!7}
MUA-RL~\citep{zhao2025muarl}               & Reward specification    & LLM-simulated users are integrated into the RL loop, so task success depends jointly on communication and tool-use behavior. \\
\rowcolor{cAgent!2}
UI-R1~\citep{lu2025uir1}                   & Reward specification    & A rule-based action reward verifies the predicted action type and coordinates of a GUI step. \\
\rowcolor{cAgent!7}
UI-AGILE~\citep{lian2025uiagile}           & Reward specification    & A continuous grounding reward with crop resampling relieves the sparse binary GUI reward. \\
\rowcolor{cAgent!2}
MAGRPO~\citep{liu2025magrpo}               & Multi-agent (boundary)  & The group-relative advantage of GRPO is generalized to cooperative multi-agent RL. \\
\bottomrule
\end{tabularx}
\end{table}

\paragraph{Reward side.}
\textbf{How do reward-side failures from Part~II persist in Agentic RL?}
Part~II identifies reward-side limitations at two levels: the reward signal $R_i$ itself, and the advantage $\hat{A}_i$ that turns it into a per-token weight.
At the reward-signal level, failures arise in reward density, reward source, reward shaping, and multi-objective aggregation.
At the advantage level, failures arise from length and variance normalization bias in $\hat{A}_i$, degenerate baselines when all group responses receive the same reward, and coarse credit assignment that broadcasts the advantage uniformly across all token positions.
These limitations persist in the agentic setting. As on the trajectory side, the methods below are illustrative. Each shows a Part~II reward-side failure reappearing in the agentic setting rather than offering a complete catalog.

On reward source, open-ended environments require task-specific verifier designs. Search-R1~\citep{jin2025search} uses exact match over the final answer, SWE-RL~\citep{wei2026swe} uses a lightweight rule-based reward, such as a similarity score between the ground-truth and LLM-generated solutions, as a graded terminal signal, and ReTool~\citep{feng2025retool} derives the reward from whether tool calls ultimately produce correct outcomes, rewarding successful tool-use strategies without requiring step-level annotation.
ARTIST~\citep{singh2025artist} interleaves reasoning, tool queries, and tool outputs in one trajectory and lets an outcome reward decide when and which tool to invoke. It couples a trajectory-side change, the interleaving of action and observation tokens, with a reward-side choice of outcome-only credit over a heterogeneous action sequence.
On degenerate baselines, filtering prompts by difficulty no longer stabilizes $\sigma_G$. In the agentic setting task difficulty depends on the tool-call sequence the policy chooses, not on the prompt alone, so it is harder to ensure that each group contains at least one successful trajectory.

\textbf{What new reward-side failures are introduced by multi-turn interaction with the environment?}
Multi-turn interaction introduces two new reward-side problems whose severity grows with the horizon $H$ and environment complexity.

\emph{Temporal credit assignment}: a terminal reward delivered after $H$ turns must be attributed to tool calls made many steps earlier, and uniformly broadcasting $\hat{A}_i$ across all turns discards all intermediate evidence.
SWEET-RL~\citep{zhou2025sweetrl} addresses this problem by training a turn-level critic with access to privileged training-time information that is unavailable to the actor, producing step-wise evaluative signals that reduce the need to broadcast a single terminal reward uniformly across all turns.
MT-GRPO~\citep{wei2025mtgrpo} keeps the critic-free structure and adds turn-level rewards, blending each turn's immediate feedback with a share of the final advantage so early turns are not credited by the terminal reward alone. IGPO~\citep{wang2025igpo} instead rewards the information gain of each turn, read from the model's own belief update, giving dense supervision without an external reward model. Both refine how $\hat{A}_i$ is spread across turns rather than broadcast uniformly.

\emph{Reward specification}: open-ended environments do not admit a single binary verifier, so the reward must be designed to match the natural feedback structure of each environment.
WebRL~\citep{qi2024webrl} uses an outcome-supervised reward model to score whether the final web-task outcome is successful, and integrates this signal into a self-evolving curriculum for online agent training.
MUA-RL~\citep{zhao2025muarl} instead emphasizes dynamic multi-turn user interaction by integrating LLM-simulated users into the RL loop, illustrating a reward-specification setting in which task success depends on both communication and tool-use behavior.
GUI control makes reward specification concrete. UI-R1~\citep{lu2025uir1} uses a rule-based action reward that checks the predicted action type and coordinates, turning a GUI step into a verifiable signal. UI-AGILE~\citep{lian2025uiagile} replaces the binary grounding reward with a continuous one and resamples crops to relieve sparse rewards. This is reward-side shaping against the degenerate baselines analyzed above.

These two dimensions are not exhaustive. They are themselves products of the same diagnostic reading, which locates new failures by asking where the environment perturbs the reward side, and the same reading keeps generating further questions.

\paragraph{Coupled failures.}
The failures above are coupled because the same loop runs through both sides. In the agentic setting these mechanisms form what we call the \emph{sparse-reward closure}: sparse terminal rewards lead to degenerate baselines, which produce uninformative gradients, which yield low-quality trajectories, which in turn make rewards even sparser.
The curriculum difficulty suitable for the current policy depends on reward statistics, which are reward-side quantities. However, these statistics depend on the trajectory distribution, which is a trajectory-side quantity.
Neither side can be fixed independently.
WebRL~\citep{qi2024webrl} can be interpreted as addressing part of this coupling by jointly adapting the curriculum, reward modeling, and online RL procedure as the policy changes.
ARPO~\citep{lu2025arpo} provides another example of this coupling: its replay buffer addresses trajectory scarcity under sparse rewards, while its task selection strategy filters for informative GUI interactions.

\paragraph{Open problems.}
The environment adds two questions that single-turn generation never faces.

\textbf{Open problem 1, the interaction mode.} An agentic rollout is a sequence of decisions interleaved with environment responses. The agent must decide when to call which tool, how to keep each call cheap, and how to read a reward from an environment with no single verifier. Existing methods answer these in isolation, and no principled account shapes the interaction protocol to keep each turn informative. The question also extends to who the agent interacts with. With several cooperating policies the environment becomes multi-agent, and reward attribution then runs across agents rather than only across turns, as in MAGRPO~\citep{liu2025magrpo}.

\textbf{Open problem 2, the update mode.} Updating on a collected multi-turn trajectory couples both sides. On the trajectory side, the state distribution drifts across turns, so importance ratios under $\piold$ go stale, and single-turn fixes such as DAPO clipping or the GSPO sequence ratio do not transfer to interleaved trajectories. On the reward side, a terminal reward must be spread across many earlier turns, yet temporal credit and reward specification are not designed jointly. Together these form the sparse-reward closure above. No method yet derives from $J(\theta)$ when the two sides must co-adapt, so a principled multi-turn update rule remains open.

\paragraph{Summary.}
Applying the framework to Agentic RL exercises its three properties at once.
\emph{First-principles}: the agentic objective is still $J(\theta)=\E[R(\tau)]$, now over multi-turn trajectories, so no new starting point is needed.
\emph{Diagnostic}: every inherited failure keeps its trajectory-side or reward-side location, and each new failure the environment introduces lands on the same two axes, namely rollout validity, importance ratio staleness, and interaction cost on the trajectory side, and temporal credit assignment and reward specification on the reward side. The same reading also exposes the sparse-reward closure, a coupled failure in which sparse rewards and degenerate trajectories feed each other and neither side can be fixed alone.
\emph{Extensible}: the two axes carry over to a structurally harder regime without a new organizing principle, which is the main analytical contribution of this section.

\section{On-Policy Distillation Combined with GRPO}
\label{sec:opd_line}

\subsection{Original OPD: Replacing the Reward Objective}
\label{sec:opd_original}

On-Policy Distillation (OPD) in its original form abandons the reward-maximization objective entirely.
Rather than optimizing $J(\theta) = \E_{\tau\sim\pitheta}[R(\tau)]$, the student is trained to minimize a divergence from a teacher policy $\pi_T$ on trajectories sampled from the student's own distribution.
The two canonical objectives are the reverse KL, used by MiniLLM~\citep{gu2024minillm},
\begin{equation}
  \label{eq:opd_rkl}
  \mathcal{L}_{\mathrm{RKL}}(\theta)
  = \mathrm{KL}(\pitheta\|\pi_T)
  = \E_{y\sim\pitheta}\!\left[\sum_{t}\log\frac{\pitheta(y_t|x,y_{<t})}{\pi_T(y_t|x,y_{<t})}\right],
\end{equation}
which is mode-seeking and concentrates the student on high-probability teacher outputs, and the forward KL, which corresponds to the standard supervised KD objective and appears as one instantiation within GKD~\citep{agarwal2023gkd},
\begin{equation}
  \label{eq:opd_fkl}
  \mathcal{L}_{\mathrm{FKL}}(\theta)
  = \mathrm{KL}(\pi_T\|\pitheta)
  = \E_{y\sim\pi_T}\!\left[\sum_{t}\log\frac{\pi_T(y_t|x,y_{<t})}{\pitheta(y_t|x,y_{<t})}\right],
\end{equation}
which is mode-covering and encourages the student to support all teacher modes.
Both objectives replace $R(\tau)$ with a token-level divergence, eliminating the verifier entirely.
This yields a distinct and well-studied line of literature focused on efficient knowledge transfer from stronger to weaker models~\citep{gu2024minillm,agarwal2023gkd}.

\subsection{This Survey's Scope: Teacher Signal Inside a Reward Objective}
\label{sec:opd_scope}

This section does not survey the independent OPD line.
Throughout this survey the optimization objective remains reward maximization: $J(\theta) = \E_{\tau\sim\pitheta}[R(\tau)]$.
The methods covered here retain this objective and ask a different question: \emph{when a teacher $\pi_T$ is available alongside a verifier, how should the teacher's per-token signal be incorporated into the GRPO update without abandoning the reward-maximization framework?}
We call this construction the \textbf{GRPO-OPD hybrid}, a teacher signal placed inside the GRPO objective rather than a replacement for it. The term recurs throughout this survey to separate this construction from the original OPD line above.

The teacher contributes through a dense per-token signal
\begin{equation}
  \label{eq:teacher_signal}
  \tilde{r}_t \;=\; d\!\left(\pi_T(\cdot|x,y_{<t}),\,\pitheta(\cdot|x,y_{<t})\right),
\end{equation}
where $d(\cdot,\cdot)$ measures the per-token alignment between the teacher and student distributions at position $t$, a quantity that reduces to a divergence in expectation~\citep{hybrid_rl_il2025}.
A common choice is the teacher-preference log-ratio
\[
  \tilde{r}_t =
  \log\frac{\pi_T(y_t|x,y_{<t})}{\pi_\theta(y_t|x,y_{<t})}.
\]
Under student-generated rollouts, maximizing this quantity corresponds to minimizing the student-to-teacher reverse KL on the visited prefixes, since its expectation equals
\(-\mathrm{KL}(\pi_\theta\|\pi_T)\). When written as a minimization loss, the corresponding reverse-KL contribution appears with the opposite sign, \(\log(\pi_\theta/\pi_T)\).
Unlike the scalar verifier reward $R_i$, $\tilde{r}_t$ is defined at every token position in every rollout.

\subsection{A Unified Formulation}
\label{sec:opd_unified}

To locate where $\tilde{r}_t$ can enter, recall the GRPO objective from Equation~\eqref{eq:grpo_obj}:
\begin{equation}
  \label{eq:opd_grpo_base}
  L^{\mathrm{GRPO}}(\theta)
  = \E\!\left[
      \frac{1}{G}\sum_{i=1}^{G}
      \frac{1}{|y_i|}\sum_{t=1}^{|y_i|}
      \min\!\Bigl(
        \underbrace{r_{i,t}}_{\mathcal{T}_A}\,
        \underbrace{\hat{A}_i}_{\mathcal{T}_B},\;
        \clip(r_{i,t},1-\epsilon,1+\epsilon)\,\hat{A}_i
      \Bigr)
    \right]
  - \underbrace{\beta\,\KL(\pitheta\|\piref)}_{\mathcal{T}_D},
\end{equation}
where $r_{i,t} = \pitheta(y_{i,t}|x,y_{i,<t})\,/\,\piold(y_{i,t}|x,y_{i,<t})$ and $\hat{A}_i=(R_i-\mu_G)/\sigma_G$.
The teacher $\pi_T$ can enter through four operators, either directly through its trajectories and probabilities or through the dense signal $\tilde{r}_t$:

\begin{itemize}
  \item $\mathcal{T}_A$: the \textbf{importance-ratio operator}, which determines the behavior policy used to generate rollouts and compute the probability ratio.
  \item $\mathcal{T}_B$: the \textbf{advantage operator}, which converts verifier scores into the per-token weight.
  \item $\mathcal{T}_C$: the \textbf{in-expectation distillation operator}, an additive on-policy distillation term inside the rollout expectation, computed on the same student-generated trajectories as the GRPO update.
  \item $\mathcal{T}_D$: the \textbf{external regularization operator}, which penalizes divergence from a reference or teacher policy outside the rollout expectation.
\end{itemize}

\begin{align}
  \label{eq:opd_unified}
  L^{\mathrm{hybrid}}(\theta)
  &= \E\!\left[
      \frac{1}{G}\sum_{i=1}^{G}
      \frac{1}{|y_i|}\sum_{t=1}^{|y_i|}
      \Bigl(
        \min\!\bigl(
          \mathcal{T}_A\,\mathcal{T}_B,\;
          \clip(\mathcal{T}_A,1-\epsilon,1+\epsilon)\,\mathcal{T}_B
        \bigr)
        + \mathcal{T}_C(\tilde{r}_t)
      \Bigr)
    \right] \notag \\
  &\quad\quad\quad\quad\quad\quad\quad\quad\quad\quad\quad\quad\quad\quad\quad\quad\quad\quad\quad\quad\quad\quad - \mathcal{T}_D(\pitheta,\pi_T,\piref),
\end{align}
where $\mathcal{T}_A = \mathcal{T}_A(r_{i,t},\pi_T)$, $\mathcal{T}_B = \mathcal{T}_B(\hat{A}_i,\tilde{r}_t)$, $\mathcal{T}_C(\tilde{r}_t)$ is an optional per-token distillation term inside the rollout expectation, and $\mathcal{T}_D$ is an external regularization term independent of the rollout.
Setting each operator to its GRPO default, $\mathcal{T}_A = r_{i,t}$, $\mathcal{T}_B = \hat{A}_i$, $\mathcal{T}_C = 0$, and $\mathcal{T}_D = \beta\,\KL(\pitheta\|\piref)$, recovers plain GRPO exactly.
Every method surveyed below is an instance of Equation~\eqref{eq:opd_unified} obtained by modifying one or more operators.

\subsection{Existing Methods: Trajectory Side and Reward Side}
\label{sec:opd_methods}

\begin{table}[!t]
\centering
\caption{GRPO-OPD hybrid methods grouped by the operator through which the teacher signal $\tilde{r}_t$ enters the GRPO objective. $\mathcal{T}_A$ is the importance ratio on the trajectory side and $\mathcal{T}_B$ is the advantage on the reward side. The final group collects coupled designs that span both sides.}
\label{tab:opd_variants}
\scriptsize
\setlength{\tabcolsep}{4pt}
\renewcommand{\arraystretch}{1.05}
\begin{tabularx}{\textwidth}{>{\raggedright\arraybackslash}p{2.4cm}>{\raggedright\arraybackslash}p{3.2cm}X}
\toprule
\rowcolor{tabheadOPD}
Method & Operator & Key mechanism \\
\midrule
\rowcolor{tabbandOPD}
\multicolumn{3}{l}{\textit{Trajectory side: teacher signal enters through $\mathcal{T}_A$, the importance ratio}} \\
\rowcolor{cOPD!8}
dGRPO~\citep{ramos2026combining} & $\mathcal{T}_A$ initialization & A cold-start SFT warm-up stabilizes the student prefix distribution before hybrid training begins. The importance ratio itself is not modified at training time. \\
\rowcolor{cOPD!2}
RLAD~\citep{rlad2025}           & $\mathcal{T}_A$ prefix quality  & A geometric ratio mixture $r^\lambda\,(r^T)^{1-\lambda}$ anchors the trust region to both the old policy and the teacher. Teacher-preferred tokens reach the clip boundary later. This permits a larger update on those tokens. \\
\rowcolor{cOPD!8}
LUFFY~\citep{yan2025luffy}      & $\mathcal{T}_A$ mixed rollouts  & Off-policy teacher traces are mixed into the rollout group and the group-relative advantage is computed over the mixed group. Regularized importance sampling amplifies low-probability crucial tokens and avoids rigid imitation. \\
\rowcolor{cOPD!2}
Prefix-RFT~\citep{huang2025prefixrft} & $\mathcal{T}_A$ mixed rollouts & A teacher prefix is spliced with a student continuation inside each trajectory. Every rollout therefore composes the teacher and student behavior distributions rather than mixing them across separate rollouts. \\
\midrule
\rowcolor{tabbandOPD}
\multicolumn{3}{l}{\textit{Reward side: teacher signal enters through $\mathcal{T}_B$, the advantage}} \\
\rowcolor{cOPD!8}
CoDistill-GRPO~\citep{codistill_grpo2025} & $\mathcal{T}_B$ reward shaping & The reward is augmented as $\tilde{R}_i = R_i + \alpha\,\overline{\tilde{r}}$. The teacher and student are co-trained jointly. Co-training addresses the teacher accessibility problem. \\
\rowcolor{cOPD!2}
KDRL~\citep{xu2025kdrl}         & $\mathcal{T}_B$ collapse        & Reward shaping causes training collapse in early stages when the teacher and verifier signals are not yet aligned. This finding motivates the $\mathcal{T}_D$ design below. \\
\midrule
\rowcolor{tabbandOPD}
\multicolumn{3}{l}{\textit{Coupled failures: teacher signal across operators (both sides)}} \\
\rowcolor{cOPD!8}
dGRPO~\citep{ramos2026combining} & $\mathcal{T}_C$ additive loss  & A per-token $\log(\pi_T/\pi_\theta)$ term is added inside the rollout expectation. It provides dense teacher supervision on student trajectories. The supervision holds independently of rollout quality. \\
\rowcolor{cOPD!2}
KDRL~\citep{xu2025kdrl}         & $\mathcal{T}_D$ external KL     & The reference-policy KL is replaced by $\beta\,D_{\mathrm{KL}}(\pi_\theta\|\pi_T)$. This term sits outside the rollout expectation, so the per-rollout reward does not filter the teacher pull. The placement avoids the collapse seen with reward shaping. \\
\rowcolor{cOPD!8}
HDPO~\citep{hdpo2025}           & $\mathcal{T}_D$ self-teacher KL & The objective adds an external term $\lambda\mathcal{L}_{\mathrm{JSD}}$ to the GRPO loss. The anchor is a privileged self-teacher conditioned on the ground truth. It engages only on cliff prompts where all rollouts fail and the advantage collapses to zero. \\
\rowcolor{cOPD!2}
RLSD~\citep{rlsd2025}           & $\mathcal{T}_A$, $\mathcal{T}_B$ decoupled & The verifier sets the update direction through $\mathrm{sign}(\hat{A}_i)$ on the advantage. A privileged self-teacher evidence ratio $w_t$ then sets the per-token magnitude on the importance ratio. Splitting the two keeps the teacher from corrupting the reward-driven direction. \\
\rowcolor{cOPD!8}
REOPOLD~\citep{reopold2025}     & $\mathcal{T}_A$, $\mathcal{T}_B$ reward-as-weight & A clipped teacher log-ratio reward $\hat{R}^{\lambda}_{i,t}$ multiplies the importance ratio directly. Reward clipping and an entropy-guided sampling schedule relax strict imitation and stabilize the resulting weight. \\
\bottomrule
\end{tabularx}
\end{table}

\paragraph{Trajectory side: modifying $\mathcal{T}_A$.}
Trajectory-side methods ask whether the teacher can replace or complement $\piold$ in the importance-ratio operator.
The central motivation is what we call \emph{prefix quality}.
When rollouts from the student are low quality, the prefix distribution $\pitheta(\cdot|x,y_{<t})$ deviates from the distribution on which the teacher was trained, which makes $\tilde{r}_t$ unreliable.
The teacher signal is informative only for prefixes that are sufficiently close to those seen during training.
We use \emph{prefix quality} as an analytical label for this mismatch rather than as an established quantity from the cited methods.
Existing methods form a spectrum ordered by how far they move the behavior distribution from $\piold$ toward $\pi_T$.

dGRPO~\citep{ramos2026combining} stays at the near end of the spectrum and addresses the cold-start form of the prefix-quality problem. Before hybrid training begins, it runs an SFT warm-up that improves the coherence of the student's long-context rollouts, so the prefixes on which $\tilde{r}_t$ is evaluated are more reliable from the first RL iteration. The importance ratio itself is not modified at training time.
RLAD~\citep{rlad2025} modifies the ratio without changing the rollout source, using a geometric mixture of $\piold$ and $\pi_T$ as the denominator:
\[
  \mathcal{T}_A(r_{i,t},\pi_T) = \left(\frac{\pitheta}{\piold}\right)^{\!\lambda}\!\left(\frac{\pitheta}{\pi_T}\right)^{\!1-\lambda},\quad \lambda\in[0,1],
\]
which anchors the trust region to both the old student policy and the teacher.
The advantage $\hat{A}_i$ remains unchanged, so the teacher affects the update only through the clipping boundary. 
With a positive advantage, teacher-preferred tokens reach the clipping threshold later, which permits a larger update. 
With a negative advantage, teacher-preferred tokens are less strongly suppressed.
Teacher guidance is therefore advantage-aware without directly modifying $\mathcal{T}_B$.

LUFFY~\citep{yan2025luffy} sits at the far end of the existing spectrum and modifies the rollout source itself: off-policy traces from a stronger teacher, such as DeepSeek-R1, are mixed into the group alongside on-policy rollouts, and the group-relative advantage is computed over the mixed group. 
A naive importance ratio assigns vanishing weight to exactly the low-probability teacher tokens that carry capabilities the student lacks, so LUFFY reshapes the weight through regularized importance sampling and amplifies the gradient contribution of these tokens while mitigating rigid imitation.
The mixed-policy estimator retains the reward-maximization objective and is accompanied by a convergence-rate analysis.
Prefix-RFT~\citep{huang2025prefixrft} carries a related mixing idea inside individual trajectories, splicing a teacher prefix with a student-generated continuation so that each rollout composes the two behavior distributions.

\paragraph{Reward side: modifying $\mathcal{T}_B$.}
Reward-side methods keep rollout sampling and the importance ratio unchanged, with $\mathcal{T}_A = r_{i,t}$, and introduce the teacher signal into the advantage computation.
The central motivation is that $\hat{A}_i$ is a scalar verifier reward that is broadcast uniformly to all tokens and therefore provides no per-token credit signal. The dense signal $\tilde{r}_t$ can mitigate this limitation.
In principle, $\mathcal{T}_B$ can be augmented through direct interpolation:
\[
  \mathcal{T}_B(\hat{A}_i, \tilde{r}_t) = (1-\alpha)\,\hat{A}_i + \alpha\,\tilde{r}_t,\quad \alpha\in[0,1),
\]
where $\alpha=0$ recovers GRPO.
The limit $\alpha\to 1$ would discard the verifier advantage entirely, reducing the update to pure distillation and exiting the reward-maximization frame. Accordingly, all methods within this section operate with $\alpha<1$.

In practice, however, existing methods introduce $\tilde{r}_t$ through reward shaping rather than direct advantage interpolation. The teacher log-ratio is folded into the scalar reward $R_i$ before the group-relative advantage is computed, so the GRPO normalization and clipping mechanisms remain unchanged.
CoDistill-GRPO~\citep{codistill_grpo2025} follows this design by setting $\tilde{R}_i = R_i + \alpha\cdot\overline{\tilde{r}}$ and avoids the \emph{teacher accessibility} problem by jointly co-training the teacher and the student.
The key challenge is \emph{signal conflict}. The reward $R_i$ measures outcome correctness, whereas $\tilde{r}_t$ measures token-level proximity to the teacher. These two signals can disagree when the student reaches correct answers through reasoning paths that differ from those favored by the teacher.
KDRL~\citep{xu2025kdrl} also evaluates reward shaping and finds that it causes training collapse in the early stages, which confirms that embedding the teacher signal into the advantage is unstable when the two signals are not yet aligned.

\paragraph{Coupled failures: modifying $\mathcal{T}_C$, $\mathcal{T}_D$, and beyond.}
Reward-side and trajectory-side failures are coupled.
When rollout quality is low, $\tilde{r}_t$ in $\mathcal{T}_B$ becomes noisy because the prefixes of the student deviate from the training distribution of the teacher. At the same time, $R_i$ becomes less informative because the verifier reward is sparse on difficult prompts.
The two signals can fail simultaneously, leaving the optimization process without useful gradients.
Existing methods address this issue through three strategies.

\emph{Additive loss through $\mathcal{T}_C$}: rather than modifying the advantage, these methods add the teacher signal as a separate loss term inside the rollout expectation. The term provides gradients independently of rollout quality.
dGRPO~\citep{ramos2026combining} instantiates $\mathcal{T}_C$ as a per-token reverse-KL distillation term inside the rollout expectation, with $\mathcal{T}_C(\tilde{r}_t) = \beta_d\,\tilde{r}_t^{\mathrm{RKL}}$, where $\tilde{r}_t^{\mathrm{RKL}} = -\log(\pitheta/\pi_T) = \log(\pi_T/\pitheta)$. This term is summed over tokens within the same expectation as the GRPO clipped surrogate. It thereby provides dense per-token teacher supervision on the trajectories visited by the student.

\emph{External regularization through $\mathcal{T}_D$}: these methods leave the in-expectation update alone and instead anchor the student to a teacher through a term computed outside the rollout expectation. Because the term is external, the per-rollout reward does not filter the teacher pull.
KDRL~\citep{xu2025kdrl} modifies $\mathcal{T}_D$ by replacing the reference-policy KL term with $\beta\,D_{\mathrm{KL}}(\pitheta\|\pi_T)$.
This change converts the external regularizer into a persistent pull toward the teacher. The collapse observed when the teacher signal is injected into the scalar reward is therefore avoided. KDRL is a joint KD and RL objective: the GRPO clipped surrogate maximizes verifier-based reward while the modified $\mathcal{T}_D$ simultaneously pulls the student toward the teacher.
HDPO~\citep{hdpo2025} is a second teacher-directed $\mathcal{T}_D$ method alongside KDRL. It uses a different teacher and a different trigger. Its objective is $\mathcal{L}_{\mathrm{HDPO}} = \mathcal{L}_{\mathrm{GRPO}} + \lambda\,\mathcal{L}_{\mathrm{JSD}}$, where $\mathcal{L}_{\mathrm{JSD}}$ is a token-averaged JSD against a privileged self-teacher on cliff prompts. The accompanying analysis frames this as a KL-regularized objective $\max_\pi \E[R(\tau)] - \beta\,\KL(\pi\|\pitheta(\cdot|x,g))$. Here the reference anchor is replaced by a teacher conditioned on the ground-truth $g$. This is exactly the $\mathcal{T}_D$ substitution that KDRL performs.
What distinguishes HDPO from KDRL is the construction of $\pi_T$ and the condition under which the term is active. The teacher is a prompt-conditioned variant of the student rather than a separate model. The JSD anchor engages only on the all-fail cliff groups where the GRPO advantage has collapsed to zero. This points to a broader class of $\mathcal{T}_D$ methods. Such methods vary the source, scope, and trigger of the teacher anchor rather than the operator through which it enters.

\emph{Coupling $\mathcal{T}_A$ and $\mathcal{T}_B$}: these methods keep the teacher inside the GRPO update but split its influence across the importance ratio and the advantage. The two operators are driven by different signals so that the teacher never overrides the verifier.
RLSD~\citep{rlsd2025} is a self-distilled RLVR method. It splits the per-token update across $\mathcal{T}_A$ and $\mathcal{T}_B$ rather than folding the teacher into a single scalar. The method first constructs a privileged self-teacher conditioned on the ground-truth answer. It then separates the update into a direction and a magnitude, and it sources the two from different operators.
The verifier reward determines the update direction through $\mathrm{sign}(\hat{A}_i)$, which is a quantity on the advantage operator $\mathcal{T}_B$. The privileged self-teacher log-ratio then determines the per-token magnitude through $w_t = (\pi_T/\pitheta)^{\mathrm{sign}(\hat{A}_i)}$. This magnitude reweights the importance-ratio term $\mathcal{T}_A$. The teacher sets only the magnitude, and it never corrupts the reward-driven direction.
REOPOLD~\citep{reopold2025} treats the teacher--student log-likelihood ratio as a token-level reward. This reward is multiplied directly onto the importance ratio. The objective takes the form $\rho_{i,t}(\theta)\,\hat{R}^{\lambda}_{i,t}(\theta)\,M^{(k)}_{i,t}$, where $\hat{R}^{\lambda}_{i,t}$ is a clipped teacher reward and $M^{(k)}_{i,t}$ is a token-level mask. The teacher signal therefore enters as a reward weight on $\mathcal{T}_B$ that acts through the $\mathcal{T}_A$ ratio. It is not a separate additive $\mathcal{T}_C$ term as in dGRPO.
The method relaxes the strict imitation constraint of standard on-policy distillation. It stabilizes the resulting reward through three mechanisms. These are mixture-based reward clipping, entropy-guided token-level dynamic sampling, and an exploration-to-refinement training schedule.

\subsection{Summary and Open Problems}
\label{sec:opd_summary}

\paragraph{Summary.}
Combining OPD with GRPO exercises the three properties of the framework at once.
\emph{First-principles}: every method retains $J(\theta)=\E[R(\tau)]$ and is an instance of the unified family in Equation~\eqref{eq:opd_unified}, introducing the teacher signal $\tilde{r}_t$ through one or more of the four operators $\mathcal{T}_A$, $\mathcal{T}_B$, $\mathcal{T}_C$, $\mathcal{T}_D$.
\emph{Diagnostic}: each method is located by the operator it modifies and by the residual failure that follows. Modifying $\mathcal{T}_A$ reduces prefix-quality mismatch but strains the off-policy correction when the student and teacher are far apart. This is why LUFFY must reshape the importance weights on teacher tokens. Modifying $\mathcal{T}_B$ adds density and credit but introduces signal conflict. The update stays unstable until the teacher and verifier agree. A coupled failure then appears once rollout quality drops, because low quality degrades $\tilde{r}_t$ and $R_i$ at the same time. The teacher must then enter at the level of the full objective. dGRPO adds it inside the rollout expectation through $\mathcal{T}_C$. KDRL and HDPO instead place a teacher-directed term in $\mathcal{T}_D$ outside the expectation. RLSD and REOPOLD take a third route and couple $\mathcal{T}_A$ with $\mathcal{T}_B$. RLSD lets the verifier set the update direction while the teacher sets only the per-token magnitude. REOPOLD multiplies a clipped teacher reward directly onto the importance ratio.
\emph{Extensible}: the same two axes carry over. The reward side is now driven by a dense teacher signal rather than a verifier scalar. This shift is what makes the four-operator account a reward-side structural extension rather than a new framework.

\paragraph{Open problems.}
The four-operator structure of Equation~\eqref{eq:opd_unified} exposes one gap per operator: existing substitutions cluster near the GRPO default, and the principled version of each is missing.
\textbf{Gap 1, the substitution endpoint of $\mathcal{T}_A$}: existing methods move the behavior distribution only partway toward $\pi_T$. The limiting configuration in which $\pi_T$ fully replaces $\piold$ as the sole behavior policy remains unexplored to our knowledge. The framework makes the obstacles visible. The importance correction degrades as the student and teacher grow further apart. The group-relative baseline also degenerates when near-uniformly successful teacher rollouts drive $\sigma_G \to 0$.
\textbf{Gap 2, the teacher signal as a reward in $\mathcal{T}_B$}: existing methods route the teacher into the advantage through heuristic reward shaping, as in CoDistill-GRPO, which folds the teacher log-ratio into the scalar reward before normalization. None of these designs derives the optimal mixing between $\hat{A}_i$ and $\tilde{r}_t$ from the structure of $J(\theta)$. The deeper gap is that $\tilde{r}_t$ is itself a reward signal, yet none of the Part~II reward-side machinery has been applied to it. Whether it should carry its own group-relative baseline, credit assignment, or bias correction is open.
\textbf{Gap 3, the in-expectation term of $\mathcal{T}_C$}: dGRPO fixes this term to a reverse KL with a constant weight. The choice of divergence, the weight schedule, and the interaction with the clipped surrogate all remain unexplored. The operator is occupied by a single method. Even basic alternatives such as a forward-KL or JSD in-expectation term remain untried.
\textbf{Gap 4, the external regularizer of $\mathcal{T}_D$}: KDRL replaces the reference KL with a teacher KL wholesale. 
HDPO swaps in a privileged self-teacher KL that is gated to cliff prompts. In both methods the choice of anchor, KL direction, coefficient schedule, and trigger remains heuristic. 
There is also no principled account of when the teacher pull should enter inside the rollout expectation through $\mathcal{T}_C$ and when it should act outside it through $\mathcal{T}_D$. 
The contrast between KDRL and HDPO already shows that the placement and the gating change stability.

Read from first principles, these gaps share one root question: how should a teacher reward be introduced to guide the optimization objective without abandoning $J(\theta)$. 
Two directions follow from this question. The first is to perfect each operator on its own. 
Every operator above is still occupied by a handful of heuristic instantiations, so the principled form of $\mathcal{T}_A$, $\mathcal{T}_B$, $\mathcal{T}_C$, and $\mathcal{T}_D$ each remains to be derived from the structure of $J(\theta)$. 
The second direction is to fuse operators rather than treat them in isolation. The four gaps do not act independently. 
Most methods already pair one primary operator with auxiliary choices about initialization, scheduling, or teacher construction, and RLSD and REOPOLD show that a single method can span $\mathcal{T}_A$ and $\mathcal{T}_B$ at once. 
A systematic account of how the teacher signal should be split across operators, and of which operator combinations are complementary rather than redundant, is therefore a promising direction that remains underexplored.

\section{Discussion}
\label{sec:discussion}

\subsection{Scope and Framework Boundary}
\label{sec:alternatives}

This survey focuses on the policy gradient line, in which the optimization objective remains reward maximization: $J(\theta) = \E_{\tau \sim \pitheta}[R(\tau)]$.
Before drawing conclusions from the two-sided framework, it is necessary to situate two prominent families of methods that address similar infrastructure costs from a structurally different direction.

\paragraph{Direct Preference Optimization.}
DPO~\citep{rafailov2023direct} begins from the same KL-regularized RLHF objective as PPO but observes that its optimality condition admits a closed-form solution in terms of the policy itself.
Substituting this solution into a preference likelihood yields a supervised classification objective that eliminates the reward model, online rollouts, and critic entirely.
The trajectory probability side disappears not because it was improved but because the new objective performs no rollout.
This is not a refinement of the estimator: it replaces the optimization problem entirely.
From the two-sided perspective, DPO is neither a trajectory-side nor a reward-side fix. It exits the policy gradient frame.
What DPO gains is simplicity and training stability. What it loses is the ability to perform on-policy exploration, which is precisely what makes GRPO-line methods effective on verifiable tasks where a reward signal is available at inference time.

\paragraph{Divergence-minimization On-Policy Distillation.}
The independent OPD line, represented by MiniLLM~\citep{gu2024minillm} and GKD~\citep{agarwal2023gkd}, minimizes a divergence from a teacher on student-generated trajectories but does not retain $J(\theta)$ as the objective.
The gradient of $\mathrm{KL}(\pitheta \| \pi_T)$ points toward increasing similarity to the teacher, not toward increasing task reward.
Although the trajectory probability factor is present because trajectories are sampled from $\pitheta$, the reward factor is replaced by a divergence that carries no notion of task performance.
As a consequence, these divergence-minimization OPD methods cannot benefit from process reward supervision, group-relative baselines, or any of the reward-side improvements in Part~II, because none of those constructs applies to a pure divergence objective.
The GRPO-OPD hybrid surveyed in Section~\ref{sec:opd_line} is structurally distinct: it retains $J(\theta)$ as the primary objective and uses the teacher signal as one component of the reward, not as a replacement for it.

Methods that exit this frame, whether by eliminating rollouts as DPO does or by replacing the reward factor with a divergence as in the divergence-minimization OPD line, require their own first-principles derivations, which proceed from different starting objectives and produce different gradient directions.
Comparing methods across this boundary by empirical benchmark alone, without accounting for the structural difference in what each gradient actually optimizes, is a common source of misleading conclusions in the literature.

\subsection{The Two-Sided Framework: Unified, Diagnostic, and Extensible}
\label{sec:chain_summary}

The introduction identifies three properties the framework must have: it must be unified and first-principles, diagnostic, and extensible.
Applying the framework across all three parts confirms that each property holds in practice.

\textbf{Unified and first-principles.}
Every method reviewed, from REINFORCE to GRPO to the latest agentic and distillation variants, can be located on the two-sided map as a modification of one or both factors of $\nabla_\theta J(\theta)$ in response to a specific failure the preceding formulation left unresolved.
This traceability is not merely organizational.
It reveals that what appears in the literature as a collection of independent proposals is in fact a structured sequence of failure-motivated responses along two axes.

\textbf{Diagnostic.}
When a method fails, whether through entropy collapse, reward over-optimization, exposure bias, or sparse training signal, the failure can be located on the map before a fix is proposed. Reward hacking, for instance, is located on the reward side as a failure of signal fidelity and bounded by the KL penalty toward $\piref$ (Section~\ref{sec:reward_source}).
A trajectory-side failure such as ratio drift or rollout diversity collapse cannot be resolved by a reward-side intervention, and vice versa.
Many published methods that appear to address compound failures are modifying only one side while leaving the other unresolved.
The map makes this visible and prevents the confusion of symptom with cause.
Given any new method, three questions apply: which factor does it modify, in response to which failure, and what residual instability does the modification introduce.

\textbf{Extensible.}
The same two axes apply beyond single-turn GRPO to both structural extensions in Part~III.
The primary differentiator across the three regimes is the nature of the reward signal entering the advantage estimate.
The GRPO line uses a verifiable outcome scalar normalized by group statistics.
The Agentic extension uses an environment return accumulated over multi-turn interaction.
The GRPO-OPD hybrid uses a dense per-token teacher log-probability.
The advantage estimation layer inherits a different failure structure in each case, yet the two-sided decomposition applies identically to all three.

Compound failures confirm this: Part~II identifies clip--variance coupling and granularity coupling as cases where single-side fixes leave residual instability on the other side.
Part~III adds the sparse-reward closure in the agentic setting and the simultaneous degradation of verifier reward and teacher signal in the GRPO-OPD hybrid setting.
All four are located on the same two axes.
The open problems that appear throughout Parts~II and~III share the same structure: they are termination points of the first-principles derivation where neither side has yet found a principled resolution, not gaps in coverage.

\subsection{Open Problems at the Intersection of Lines}
\label{sec:open_cross}

Parts~II and~III each close with the open problems internal to their own regime: the both-side gaps of single-turn GRPO in Section~\ref{sec:both_variants}, the interaction-mode and update-mode problems of the agentic setting in Section~\ref{sec:agentic}, and the four operator-level gaps of the GRPO-OPD hybrid in Section~\ref{sec:opd_summary}.
The open problems collected here are not new entries in that list. They are the same missing joint theory, recurring one level up.

\paragraph{One missing theory across three scales.}
Section~\ref{sec:both_variants} states the root gap for the single-turn line: $J(\theta)$ couples $p_\theta(\tau)$ with $R(\tau)$, yet no theory derives when a trajectory-side and a reward-side fix must co-vary rather than act alone.
The same gap reappears at two larger scales once the two structural extensions are admitted.
At the line-to-line scale, the question is whether a fix established on one axis in one regime transfers to the same axis in another regime.
At the combination scale, it is whether fixes drawn from different lines compose into a well-behaved joint system, the cross-line form of the warning in Section~\ref{sec:both_variants} that independently validated single-side methods carry no such guarantee.
Each scale is the previous one with more axes free to interact, and none has a principled account.
The first systems to combine both Part~III extensions make the combination scale concrete.
When a teacher signal is placed inside a multi-turn agentic objective, the naive combination of on-policy distillation and GRPO destabilizes training, and stability is recovered only by gating the teacher term or by a curriculum on trajectory depth~\citep{lu2026sdar,wang2026tcod}.
This is an emerging direction rather than a settled one, and no theory yet predicts when the two lines compose safely.

\paragraph{Transfer across lines.}
The transfer question is sharp because the single-turn fixes are defined for fixed-length single-turn responses, so their validity in a multi-turn or teacher-driven regime cannot be assumed.
Toward the agentic extension, asymmetric clipping from DAPO is motivated by single-turn entropy collapse, and whether the same mechanism operates at the same rate across multi-turn trajectories with interleaved observations is open. The sequence-level ratio of GSPO is defined for a fixed-length response, and extending it to action--observation interleaving requires a non-trivial reformulation.
Toward the GRPO-OPD hybrid, Gap~2 of Section~\ref{sec:opd_summary} is exactly this transfer on the reward side: the teacher signal $\tilde{r}_t$ is itself a reward signal, yet the Part~II density, credit-assignment, and bias-correction interventions have not been applied to it.
In every case the framework locates the question precisely, namely which single-turn fix, on which axis, transfers to which extension, while the empirical and theoretical study remains to be done.

\subsection{Extensions Beyond the Single-Model, Single-Modality Setting}
\label{sec:extensions}

The survey focuses on single-model, text-only policy optimization.
Two extensions are worth noting as directions where the two-sided framework applies but requires adaptation.

\paragraph{Diffusion language models.}
Diffusion language models~\citep{nie2025llada,sahoo2024mdlm} generate text by iterative denoising rather than autoregressive token prediction.
The trajectory probability factor $\log \pitheta(\tau)$ in the policy gradient estimator is then defined over a sequence of denoising steps rather than token selections.
The importance ratio and clipping mechanisms developed for autoregressive models do not transfer directly: the denoising trajectory has a different Markov structure, and the per-step action space is a distribution over masked tokens rather than a vocabulary distribution.
Early work~\citep{zhao2025d1,rojas2025gdpo} applies GRPO-style updates to diffusion LLMs by treating each denoising step as an action, but the reward-side interventions, including process reward and group-relative baseline, are not yet adapted to the denoising setting, where intermediate states have a different interpretation than intermediate reasoning steps.
Whether the two-sided decomposition remains the right organizing principle for diffusion LLM policy optimization, and how the trajectory probability factor should be re-derived for the denoising Markov chain, is an open structural question.

\paragraph{Multimodal language models.}
Multimodal language models~\citep{liu2023llava,chen2024internvl} extend the action space to include both text tokens and visual inputs processed through a vision encoder.
The trajectory probability side extends relatively naturally: autoregressive text generation proceeds as in the text-only case, and the vision encoder parameters can be treated as part of the policy or held fixed.
The reward side is more complex.
Verifiable reward signals that work well for text, such as math correctness and code execution, do not cover the full range of multimodal tasks, and the LLM-as-judge reward source must be adapted to evaluate outputs that depend jointly on visual and textual inputs.

Recent work~\citep{huang2025visionr1,r1sharevl2025} applies GRPO-line methods to multimodal models, but the reward-side taxonomy developed in Part~II is not yet fully applied: multi-objective balancing across text and vision objectives, process reward supervision for visual reasoning chains, and advantage granularity for vision-language interleaved generation remain open engineering problems.
The two-sided framework provides a clear agenda: systematically apply each reward-side and trajectory-side intervention from Part~II to the multimodal setting, identify which transfer and which require adaptation, and resolve the failures that are new to the multimodal regime.

\section{Conclusion}
\label{sec:conclusion}

This survey revisited LLM policy optimization from the shared objective $J(\theta) = \E[R(\tau)]$. 
The objective has two factors, the trajectory probability and the reward, and these define the two axes along which every method is placed. 
Read this way, the rapid post-GRPO expansion is not a collection of independent proposals but a structured response to failures that arise, in a predictable order, along the trajectory and reward axes. 
The framework is therefore diagnostic: each method is located by which factor it modifies, what failure motivates the modification, and what residual instability remains. 
It is also extensible, since the same two axes carry from single-turn GRPO to the agentic and distillation extensions with no new organizing principle. 
Methods that replace $J(\theta)$ itself, such as DPO and divergence-minimization distillation, fall outside this frame as boundary cases, so comparisons across it require accounting for the structural difference rather than benchmark numbers alone.

The open problems in Section~\ref{sec:open_cross} are the precise locations where the derivation stops generating solutions. 
They recur as one missing joint theory across three scales, from the two sides of a single update to the transfer and combination of fixes across lines. 
This is the most actionable output of a first-principles reading: it tells a researcher not only what has been solved but where the next solution must come from.

\clearpage
\bibliography{references}

\end{document}